\documentclass[lettersize,journal]{IEEEtran}
\usepackage{amsmath,amsfonts}
\usepackage[linesnumbered,ruled,vlined]{algorithm2e}
\usepackage{array}
\usepackage[caption=false,font=footnotesize,labelfont=sf,textfont=sf]{subfig}
\usepackage{textcomp}
\usepackage{stfloats}
\usepackage{url}
\usepackage{verbatim}
\usepackage{graphicx}
\usepackage{cite}
\usepackage[table]{xcolor} 
\usepackage[hidelinks]{hyperref}
\hypersetup{
	colorlinks=true,
	linkcolor=blue,
	citecolor=blue,
	filecolor=magenta,
	urlcolor=black
}
\usepackage{float}
\usepackage{color}
\usepackage{threeparttable}
\usepackage{booktabs}
\usepackage{bbding}
\usepackage{multirow}
\usepackage{makecell}
\definecolor{sred}{RGB}{255,127,80}
\definecolor{red}{RGB}{255,0,0}
\usepackage{booktabs} 
\usepackage{svg}
\usepackage{soul}
\usepackage{amssymb}
\usepackage{eso-pic}
\usepackage{xcolor}
\usepackage{newtxtext,newtxmath} 

\newcommand\WatermarkText{%
	This work has been submitted to the IEEE Transactions on Circuits and Systems for Video Technology. Copyright may be transferred
	without notice, after which this version may no longer be accessible. ©IEEE
}

\AddToShipoutPictureBG*{%
	\AtPageUpperLeft{%
		\setlength{\unitlength}{1pt}%
		\put(0,-38){
			\makebox[\paperwidth]{%
				\centering
				\parbox{\textwidth}{\centering\textcolor[gray]{0.5}{\rmfamily\small\WatermarkText}} 
			}%
		}
	}
}

\begin{document}
	\title{A Label-Free and Non-Monotonic Metric for Evaluating Denoising in Event Cameras}
	\author{Chenyang Shi, Shasha Guo, Boyi Wei, Hanxiao Liu, Yibo Zhang, Ningfang Song, Jing Jin\IEEEauthorrefmark{1}\thanks{Chenyang Shi, Boyi Wei, Yibo Zhang, Ningfang Song, Jing Jin are with the School of Instrumentation and Optoelectronic Engineering, Beihang University, Beijing, 100191, China, and also with the Tianmushan Laboratory, Hangzhou, 311115, China (e-mail): shicy@buaa.edu.cn, jinjing@buaa.edu.cn.}\thanks{Shasha Guo is with the College of Electronic Engineering, National University of Defense Technology, Hefei, 230031, China.}
		\thanks{Hanxiao Liu is with the Department of Precision Instrument, Tsinghua University, Beijing, 100080, China.}
		\thanks{\IEEEauthorrefmark{1} Corresponding author.}
	}
	\IEEEpubidadjcol
	
	\maketitle
	\begin{abstract}
		Event cameras are renowned for their high efficiency due to outputting a sparse, asynchronous stream of events. However, they are plagued by noisy events, especially in low light conditions. Denoising is an essential task for event cameras, but evaluating denoising performance is challenging. Label-dependent denoising metrics involve artificially adding noise to clean sequences, complicating evaluations. Moreover, the majority of these metrics are monotonic, which can inflate scores by removing substantial noise and valid events. To overcome these limitations, we propose the first label-free and non-monotonic evaluation metric, the area of the continuous contrast curve (AOCC), which utilizes the area enclosed by event frame contrast curves across different time intervals. This metric is inspired by how events capture the edge contours of scenes or objects with high temporal resolution. An effective denoising method removes noise without eliminating these edge-contour events, thus preserving the contrast of event frames. Consequently, contrast across various time ranges serves as a metric to assess denoising effectiveness. As the time interval lengthens, the curve will initially rise and then fall. The proposed metric is validated through both theoretical and experimental evidence.

	\end{abstract}
	

	\section{Introduction}
	\IEEEPARstart{E}{vent} cameras\cite{lichtensteiner2008vision,zhu2021flexible,zhou2023full,schoepe2024finding}, also known as dynamic vision sensors (DVS), are inspired by biological vision system\cite{medathati2016bio} and simulate the imaging mechanism of retinas. They encode relative changes in light intensity instead of absolute values, providing high temporal resolution, high dynamic range, and low power consumption. Consequently, their output is an event stream rather than frames with a fixed frame rate. Event cameras boast microsecond-level temporal resolution, allowing them to accurately and thoroughly capture rotations and high-speed movements, tasks that pose significant challenges for traditional cameras.
	
	However, due to their extreme sensitivity to changes in light intensity, event cameras trigger a substantial amount of noise in low light conditions, including background activity (BA) noise\cite{lichtensteiner2008vision}, hot noise\cite{graca2021unraveling}, and $1/f$ noise \cite{graca2021unraveling} (also known as shot noise). The presence of these noise increases the output load and severely impairs the quality and efficiency of downstream tasks. These tasks include image reconstruction\cite{wang2024asynchronous,10258440}, super-resolution\cite{wang2020joint,9625006}, motion analysis\cite{nunes2021robust,liu2023motion}, and pose estimation\cite{zhou2021event,zuo2024cross}. Consequently, the inherent noise challenges the advantages of event camera, such as high efficiency and low power consumption. Thus, denoising emerges as a fundamental and crucial task for event cameras. 
	
	Many denoising methods \cite{wu2020probabilistic,baldwin2020event,duan2021eventzoom,9241010, guo2022low,duan2022guided,duan2023neurozoom,rios-navarro2023within,mohamed2022dba} for event camera have been proposed. However, no universally accepted standard exists for evaluating the performance of denoising methods. The most commonly used approach involves using simulated event generation methods to create noise-free event sequences, followed by the addition of labeled noise. Evaluation is then performed by counting the number of valid events and noise events before and after denoising. However, artificially adding noise is not ideal, as the method of converting data differs from actual events. Additionally, evaluating added noise is troublesome since the data stream obtained by the event camera inherently contains noise. Consequently, some methods that do not rely on labels have also been proposed.
	
	Label-free denoising evaluation methods for event sequences, like EventZoom\cite{duan2021eventzoom}, use indirect approaches by estimating their effect on downstream tasks, such as super-resolution performance, to assess effectiveness. Moreover, some direct and label-free denoising evaluation metrics have been introduced, such as event structural ratio (ESR)\cite{ding2023e}. Unfortunately, ESR is a monotonic metric. In other words, the higher the amount of noise and valid events removed, the better the ESR score, which is contrary to the denoising task's objective of removing as much noise as possible while retaining more valid signals.
	
	In this paper, we report the first label-free and non-monotonic evaluation metric, area of continuous contrast curve (AOCC), for event camera denoising performance. It is evaluated based on the area under the continuous contrast curve (CCC). Our main contributions are summarized as follows:
	\begin{itemize}
		\item We propose the first label-free and non-monotonic evaluation metric, area of continuous contrast curve (AOCC), for denoising performance of event camera.
		\item The proposed AOCC was compared with both label-dependent and label-free denoising evaluation methods. AOCC achieved the same objective results as the label-dependent methods without relying on labels, demonstrating its effectiveness.
		\item This metric significantly enhances the evaluation process for event camera denoising tasks, aiding in the selection of optimal parameters for denoising methods. The development of this metric enables more efficient and effective denoising solutions. 
	\end{itemize}
	
	The remainder of this article is organized as follows. First,  Section \ref{sec:back} presents the backgrounds of event camera hardware, event noise, and denoising methods. Section \ref{sec2} reviews the recent works in this field. Next, Section \ref{sec3} introduces the proposed denoising evaluation metric AOCC and it components. We then demonstrate the experimental results compared with other benchmark metrics in Section \ref{sec4}. Finally, the article is concluded with a discussion in Section \ref{sec7} and Section \ref{sec8}.
	\section{Backgrounds}
	\label{sec:back}
	In this section, we first introduce the development trends of event camera hardware and the generation model of events. Next, we describe the main types of noise encountered by event cameras. Finally, we briefly present the main denoising methods for event cameras.
	
	\subsection{Event Camera Hardware}Event cameras simulate the imaging mechanism of the human eye by realizing the three-level abstraction of the photoreceptor layer, bipolar cell layer, and ganglion cell layer in the retinal structure. They independently encode positive relative increases and negative relative decreases in light intensity, outputting events that represent these changes.
	
	Each event is represented as a tuple $e=(x,y,p,t)$, where at timestamp $t$, an event with polarity $p\in\{-1,+1\}$ is triggered at pixel $(x,y)$. A positive (ON) event represents an increase in light intensity exceeding the threshold, while a negative (OFF) event represents a decrease. Therefore, an event is triggered at a pixel as soon as the brightness change $\Delta \mathcal{L}$ reaches a given threshold. The brightness signal $\mathcal{L}$ is defined as 
	\begin{equation}
		\mathcal{L}(\mathbf{u}_k, t_k) \triangleq \log I (\mathbf{u}_k, t_k)
	\end{equation}
	where $I$ is the light intensity, $t_k$ is the timestamp, and $\mathbf{u}_k=(x_k,y_k)$ is pixel coordinate. 
	
	The ideal event generation model is established as follows.
	\begin{equation}
		\Delta \mathcal{L}(\mathbf{u}_k, t_k) \triangleq \mathcal{L}(\mathbf{u}_k, t_k) - \mathcal{L}(\mathbf{u}_k, t_k - \Delta t_k) = p_k \mathcal{C}
	\end{equation}
	where $C$ represents the threshold, $p_k$ is the polarity, and $\Delta t_k$ represents the time elapsed since the last event at the same pixel. 
	
	However, previous event cameras are merely a simplified hardware abstraction of the retina. In contrast, the retina can capture rich detailed textures, perceive colors, and keenly detect dynamics.
	
	To more closely simulate the biological visual imaging mechanism, many solutions have embedded RGB pixels into event cameras. This has resulted in sensors like asynchronous time-based image sensor (ATIS) and dynamic and active-pixel vision sensor (DAVIS) that can output both events and frames simultaneously. The latest technical trend involves using back-illuminated (BI) and stacked technology \cite{guo20233,kodama20231,niwa20232}. A DVS imaging chip now includes both high-resolution RGB pixels and event generation pixels, with event pixel resolution reaching up to one million. Additionally, the most recent solution\cite{yang2024vision} features four event generation pixel channels for red, green, blue, and white light events.
	
	However, the increase in the number of pixels and the complexity of the technology makes pixel circuits more prone to generating noise. 
	
	\subsection{Event Noise}
	As mentioned above, the noise in event cameras is mainly BA noise. BA noise refers to the noise caused by current generated by the pixels without changes in illumination. This noise is related to ambient temperature and the manufacturing process of the pixel circuit. Typical influencing factors include leakage current leading to charge accumulation \cite{suh20201280}, which exceeds the threshold and erroneously triggers events, threshold drift for triggering events, process defects generating noise, and random photon fluctuations.

	Although many new processes have been adopted to reduce the generation of BA noise, it still exists in significant amounts, particularly under low light conditions. Consequently, denoising  for event cameras has become a fundamental and critical issue to address.
	\subsection{Denoising Methods for Event Cameras}
	Current denoising methods for event cameras typically rely on the spatiotemporal correlation of events \cite{khodamoradi2018n,9241010,guo2022low}, event density \cite{wu2020probabilistic,chen2023denoising}, motion consistency of events \cite{wang2020joint,mohamed2022dba}, and learning-based approaches\cite{baldwin2020event,duan2021eventzoom, duan2022guided, duan2023neurozoom,zhang2023neuromorphic,rios-navarro2023within}.
	
	A majority  of denoising methods are based on spatiotemporal correlation. The double window filter (DWF) \cite{guo2022low} ingeniously establishes the spatiotemporal correlation of events using two first-in, first-out queues, effectively removing noise while retaining valid events.
	
	The quantified version of the multi-layer perception filter (QMLPF)\cite{rios-navarro2023within} is a learning-based denoising method that addresses denoising as a binary classification task. The QMLPF method calculates a predicted value for each event, categorizing it as noise or a valid signal. 
	
	Some denoising methods no longer treat denoising as a binary classification task, such as learning-based EventZoom\cite{duan2021eventzoom} and NeuroZoom\cite{duan2023neurozoom}. Instead, they perform projection operations on the events and integrate denoising with downstream tasks, treating it as a sub-task of the overall process. This type of denoising method prevents the direct distinction between noise and valid events, even when the event stream is pre-labeled.

	\section{Related Work}
	\label{sec2}

	Denoising methods have achieved increasingly better performance and have become more lightweight, even being efficiently implemented on embedded hardware\cite{rios-navarro2023within}. However, there is still no consensus on a standardized benchmark for evaluating these denoising methods.
	
	Currently, there are two primary methods to evaluate the denoising performance of event cameras. The first method involves using labeled event sequences, where denoising is treated as a binary classification task, allowing direct assessment with signal or noise labels. The second method employs unlabeled data for evaluation or utilizes indirect methods, evaluating denoising performance through downstream tasks that abovementioned.
	\subsection{Label-Dependent Evaluating Metrics for Denoising Methods}
	Evaluating these denoising method is straightforward when using labeled event data, where events are annotated as either noise or valid signals. Commonly, labeled data is generated by methods such as E2VID\cite{rebecq2019high} and v2e\cite{hu2021v2e}, which reconstruct events from videos to produce clean event data, subsequently introducing noise events that follow a Poisson distribution. Current metrics for assessing denoising performance with labeled data include the noise event removal rate (NeRr)\cite{xu2023denoising}, valid event removal rate (VeRr), signal-to-noise ratio (SNR), true positive rate (TPR), false positive rate (FPR), accuracy (ACC)\cite{nielsen2024data}, receiver operating characteristic (ROC)\cite{guo2022low} curve and area under the curve (AUC), etc.
	
	However, metrics such as NeRr (which is corresponding to FPR), VeRr (which is corresponding to TPR), and SNR are monotonic. For example, a high SNR might misleadingly suggest good denoising performance due to the simultaneous high rates of noise and valid signal removal, which are considered failures in denoising tasks. Using ACC, ROC and AUC provides a more comprehensive evaluation method, offering a balanced assessment of the performance of denoising methods under different parameters. However, these metrics require labeled data, which can be impractical in real-world scenarios where distinguishing between valid signals and noise is challenging. This limitation makes these metrics unsuitable for evaluating a vast number of real-scene sequences. 
	
	\subsection{Label-Free Evaluating Metrics for Denoising Methods}
	ESR\cite{ding2023e} is a recently proposed label-free metric for evaluating the performance of denoising methods. It first projects the event stream along its trajectory to a reference time, calculates the normalized total sum of squares (an indication of image contrast) of the projected event frames, and then introduces a penalty coefficient to address the issue of over-denoising. Unfortunately, this metric is monotonous. Despite the introduction of the penalty coefficient, the score increases with the removal of both effective signals and noise. The core problem lies in the initial step, the warping operation. The amount of denoising does not significantly impact the contrast of the warped image, as long as events near the reference time are not removed in large numbers.
	
	Some evaluation metrics use downstream tasks to assess denoising performance, such as DVS response probability map (DPRM)\cite{xu2023denoising}. This method necessitates the movement of the event camera and the acquisition of the camera's pose data. Other methods evaluate denoising performance by using the denoised event stream to reconstruct images and then assessing the quality of these images using metrics such as peak signal-to-noise ratio (PSNR) and structural similarity (SSIM)\cite{hore2010image}. While it is reasonable to evaluate denoising through downstream tasks, this approach introduces complexity, making the assessment of denoising performance indirect and intricate.

	\section{Proposed Methods}
	\label{sec3}
	In this section, we first present the method for calculating event frame contrast. We then introduce the definition and calculation process of the proposed continuous contrast curve and its area.
	
	\subsection{Contrast of an Event Frame}
	For an event sequence $E$ with a duration of $T$, we choose $\Delta t$ as a time interval, then the event stream will be divided into $t/\Delta t$ segments. For a time interval $t_0 \leq t \leq t_1
	$, we accumulate events within it into an event frame $I$. $I$ is defined as follows:
	\begin{equation}
		I(x, y) = 1\left(\sum_{e \in E} (t_0 \leq t \leq t_1)\right)_t
	\end{equation}
	\begin{equation}
		I = \sum I(x, y)
	\end{equation}Where $1(\cdot)$ is an indicator function, which takes the value 1 when the condition is satisfied and 0 otherwise, $I(x, y)$ is the value of $(x, y)$. We map a pixel value of 1 to a corresponding channel intensity of 255. Next, we employ the Sobel operator to compute the gradients of the event frame along both vertical and horizontal directions. Define the gradient amplitude as $G$.
	\begin{equation}
		G = \sqrt{G_x^2 + G_y^2} 
	\end{equation}
	where $G_x$ and $G_y$ are Sobel operator of vertical and horizontal directions, respectively. Contrast $C$ of an event frame can be described as:
	
	\begin{equation}
		C = \sqrt{\frac{1}{N-1} \sum_{i=1}^{N} \left(G_i - \frac{1}{N} \sum_{i=1}^{N} G_i \right)^2} 
	\end{equation}
	where $N$ is the total number of pixels in an event frame.

\begin{figure}[htbp]
		\centering
		\subfloat[RGB frame]{
			\begin{minipage}{0.48\linewidth}
				\centering
				\includegraphics[width=\linewidth]{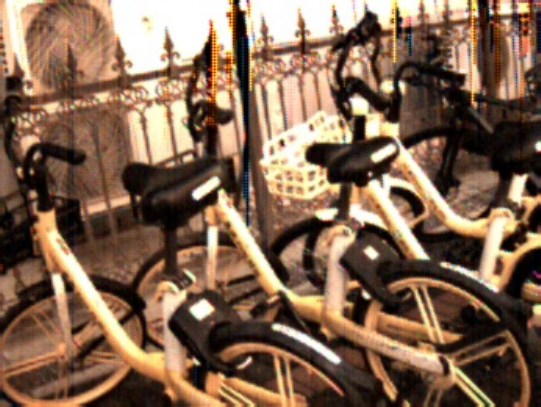}
		\end{minipage}}
		\subfloat[Grayscale frame]{
			\begin{minipage}{0.48\linewidth}
				\centering
				\includegraphics[width=\linewidth]{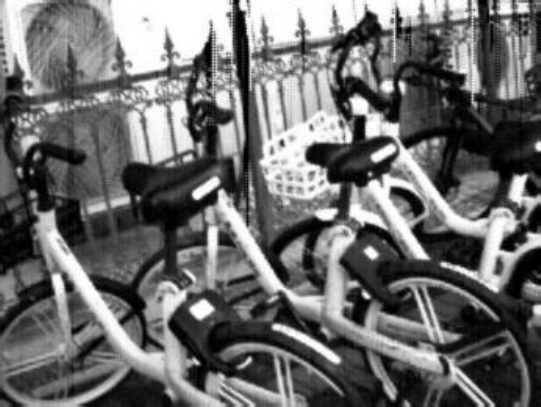}
		\end{minipage}}
		
		\subfloat[Binary frame]{
			\begin{minipage}{0.48\linewidth}
				\centering
				\includegraphics[width=\linewidth]{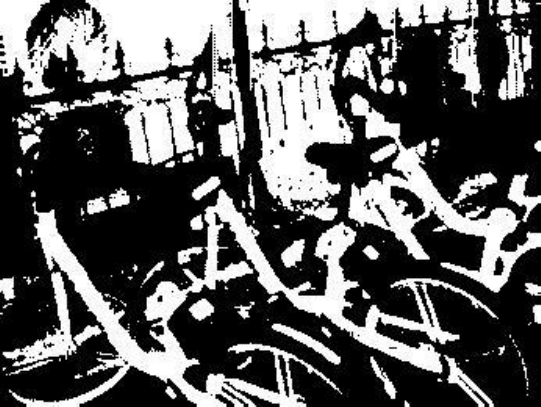}
		\end{minipage}}
		\subfloat[Event frame]{
			\begin{minipage}{0.48\linewidth}
				\centering
				\includegraphics[width=\linewidth]{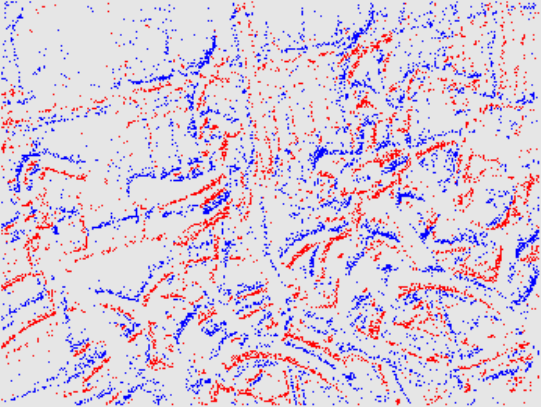}
		\end{minipage}}
		\caption{Contrast comparison of RGB frame and event frame. (a) The RGB frame captured by the active pixel sensor (APS) of a DAVIS346 event camera. (b) The grayscale image corresponding to (a). (c) The binary frame of (a) with a contrast of 211.5. (d) The event frame with a contrast of 72.9. The red pixel indicates the most recently triggered positive event, while blue indicates a negative event. The contrast of APS binary frames is significantly greater than that of event frames.}
		\label{fig:contrast}
	\end{figure}

	According to principle of event generation, events are more likely to be triggered at edge contours with significant brightness changes rather than within the scene or object, leading to a lack of fine texture information in the resulting data. Additionally, due to the sparse triggering of events, the events representing edges may not form continuous lines. Assume that, within the same brief time window, an ultra-high-speed camera and an event camera with identical fields of view operate simultaneously. In this scenario, the RGB camera will distinctly capture both the outline and more texture details of the scene, whereas the event camera will primarily record the high-brightness changes along the scene's edges. We demonstrate an example in Fig.\ref{fig:contrast}.

	\subsection{Continuous Contrast Curve}
	Let us define the contrast of an event frame \( C(t) \) as a function of time within the interval \((t, t + \Delta t)\). The event frame is represented as a binary image. If every pixel in the frame contains an event, we set all pixel values to 255. Conversely, if no events are present, all pixel values are set to 0. This scenario can be described as follows: during the accumulation period \(\Delta t\) of the event frame, each pixel either triggers an event or does not.
	
	Formally, if \(\Delta t\) is sufficiently large, \( C(t) \) approaches 0 due to the uniform presence of events across all pixels, resulting in a lack of contrast. Similarly, if \(\Delta t\) is sufficiently small, \( C(t) \) also approaches 0 because the sparse distribution of events leads to minimal contrast. Thus, \( C(t) \) is dependent on \(\Delta t\) in such a way that it is maximized for an optimal intermediate \(\Delta t\).
	\begin{equation}
		C(t) = \begin{cases} 
			0 , \text{if} \ \Delta t \to 0 \ \text{or} \ \Delta t \to \infty \\
			\text{Max} , \text{for optimal} \ \Delta t 
		\end{cases}
	\end{equation}

	As abovementioned, the contrast of an event frame is significantly lower compared to the binary frame derived from the RGB frame, a disparity stemming from the innate characteristics of sparse events. These include sensitivity to light intensity changes and the lack of color perception. 
	
	Consequently, the contrast of the RGB frame, which we denote as $C_{max}$, will surpass that of the binary event frame. For each time range, the contrast of the event frame, $C(t)$, is assumed to be less than or equal to $C_{max}$. Therefore, $C(t)$ has its upper and lower bounds. Although the specific value of the upper bound may be unknown, its existence is certain.
	
	\begin{equation}
		0 \leq C(t) \leq C_{max} 
	\end{equation}
	
	Importantly, as $C(t)$ is a continuous function, it exhibits non-monotonic behavior when $t$ is sufficiently large. This is because $C(0) = 0$ and $C(t) = 0$ for large $t$, while $C$ remains non-negative ($C \geq 0$) throughout its domain.
	
	For an event sequence $E$ with a duration of $T$. We divide it evenly into $n$ segments ($n$ is a positive integer), each corresponding to one of $n$ binary event frames. Thus, the average contrast within this duration is described as:
	
	\begin{equation}
		C_{avg} = \frac{1}{n} \sum_{i=1}^{n} C(t_i) 
	\end{equation}
	
	By selecting different values for $n$, we can determine the corresponding $C_{avg}$. Thus, $C_{avg}(t)$ can describe this process. We define $C_{avg}(t)$ as continuous-time contrast curve (CCC). The area $A_c$ of the $C_{avg}(t)$, $0 \leq t \leq \tau$, namely area of continuous contrast curve (AOCC), is described as follows.
	
	\begin{equation}
		A_c = \int_0^\tau C_{avg}(t) \, dt 
	\end{equation}
	
	To simplify calculations, we define an array of time intervals $\Delta T = \{\Delta t_0, \Delta t_1, \Delta t_2, \ldots, \Delta t_n\}$ to segment the sequence. The event sequence is duplicated $n$ times, and each duplicate is then segmented according to these predefined time intervals. As a result, each time interval division will correspond to a contrast value.
	
	\begin{equation}
		C_i \approx \frac{1}{m} \sum_{j=1}^{m} C_j, m = \left\lfloor \frac{\tau}{\Delta t_i} \right\rfloor, 0 \leq i \leq n 
	\end{equation}
	
	We then calculate the total contrast $A_c$.
	
	\begin{equation}
		A_c \approx \sum_{i=1}^{n} C_i 
	\end{equation}
	
	An ideal denoising method retains all events that delineate the edge contours of the scene or object, ensuring the highest contrast value across any accumulated time interval for event frames. By calculating the $A_c$ metric, we can assess which method delivers the best overall performance across various time intervals.
	
\begin{figure}[htbp]
		\centering
		\includegraphics[width=0.7\linewidth]{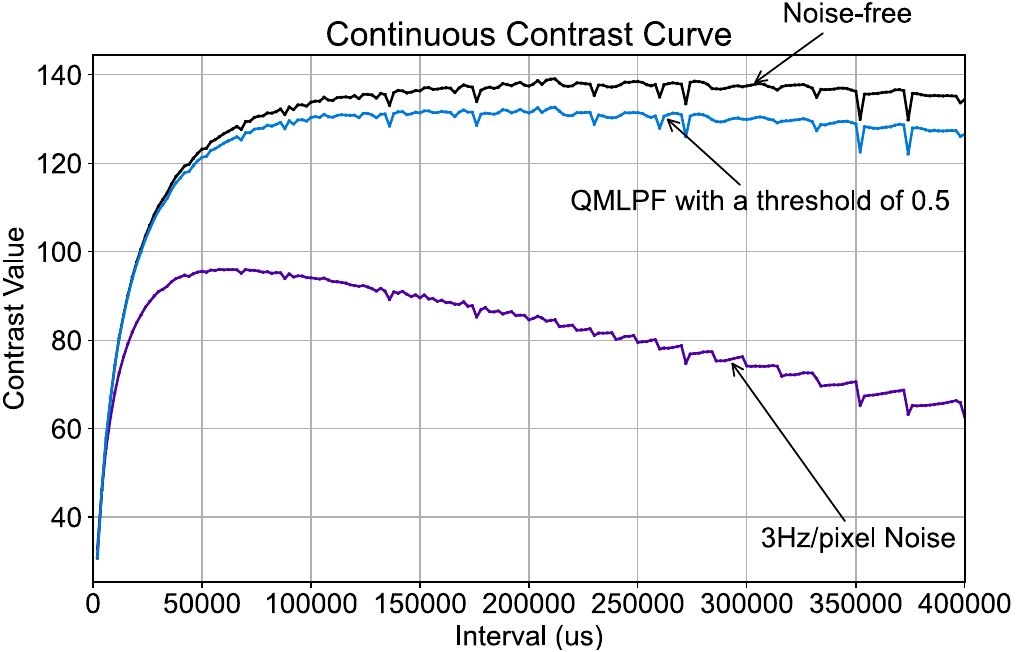}
		\caption{An example of CCC for the QMLPF method with a threshold of 0.5.}
		\label{fig:qmlpf50}
	\end{figure}

	We provide an example of drawing the CCC. We illustrate the CCC of the \textit{driving} sequence in the DND21 dataset \cite{guo2022low} without noise. Additionally, we show the event sequence with a noise level at 3 Hz/pixel. We also show the CCC of the event sequence processed by the QMLPF method with a threshold of 0.5, as depicted in Fig\ref{fig:qmlpf50}. The AOCC value can be obtained by calculating the area under the CCC. This represents the sum of the average contrast values of event frames accumulated at different time intervals.
	
	It can be observed that adding noise destructively reduces the contrast of event frames accumulated at different time intervals. After denoising, the obtained CCC follows the same trend as the CCC of the sequence without noise. This indicates that the noise is effectively separated, and the original event frame contrast is restored as much as possible.

	\section{Experimental Methodology}
	\label{sec4}
	In this section, we first introduce the experimental setup. Subsequently, we present the datasets used to evaluate denoising methods. Next, we present the existing benchmark denoising evaluation metrics. Finally, we compare the experimental results of the proposed AOCC with the benchmark metrics and provide an analysis.

	\subsection{Experimental Setup}
	We conduct two groups of experiments. In the first group, we use existing denoising methods with adjustable parameters. We first plot the CCC under different parameters. Then, we calculate the AOCC based on the CCC and other label-dependent denoising evaluation metrics and compare them. Finally, we demonstrate the advantages of using AOCC curves under different parameters to select the optimal parameters for denoising methods.

	In the second group of experiments, we test a baseline denoising method that cannot be evaluated using labels. We present the AOCC metric results for this method. Next,  the AOCC we proposed is compared with a benchmark label-free denoising metric.
	
	\subsection{Datasets} We adopt the DND21 dataset\cite{guo2022low} and the E-MLB dataset \cite{ding2023e} for evaluation. 
	
	The DND21 dataset is generated from simulated events\cite{hu2021v2e} and is inherently noise-free. We artificially introduce labeled noise at various levels to the DND21 sequences, specifically at 1 Hz/pixel, 3 Hz/pixel, and 5 Hz/pixel. The introduced noise follows a Poisson distribution. 
	
	The E-MLB dataset, recorded using the DAVIS346\footnote{[Online]. Available: https://www.inivation.cn/?list=19} event camera, comprises a total of 96 sequences captured both during the daytime and at nighttime. Each sequence exhibits a different noise level, achieved using neutral density (ND) lenses to simulate varying lighting conditions. The noise levels are categorized as ND1, ND4, ND16, and ND64, with higher values indicating greater noise levels.
	
	\subsection{Benchmark Evaluation Metrics}
	\subsubsection{Label-dependent Metrics}We use the most common label-dependent denoising evaluation metrics, noise event removal rate (NeRr), valid event removal rate (VeRr), signal-to-noise ratio (SNR), true positive rate (TPR), false positive rate (FPR), accuracy (ACC), receiver operating characteristic (ROC) curve and area under the curve (AUC) for our experiments. Their definitions are described as follows:
	
	\begin{equation}
		NeRr = \frac{TN}{TN + FP} 
	\end{equation}
	
	\begin{equation}
		VeRr = \frac{FN}{FN + TP} 
	\end{equation}
	
	\begin{equation}
		SNR = 10 \cdot \log \left(\frac{TP}{FP}\right) 
	\end{equation}
	
	\begin{equation}
		ACC = \frac{TP + TN}{TP + TN + FP + FN} 
	\end{equation}
	where True Positive (TP) classification means that a signal event is classified as a valid event, True Negative (TN) classification means that a noise event is classified as a noise event, False Positive (FP) classification means that a noise event is classified as a valid event, False Negative (FN) classification means that a signal event is classified as a noise event. The ROC curve is obtained by plotting TPR on the y-axis against FPR on the x-axis for different threshold values.
	\begin{equation}
		TPR=1-VeRr
	\end{equation}
	\begin{equation}
		FPR=1-NeRr
	\end{equation}
	\textbf{\begin{equation}
			AUC = \int_{0}^{1} TPR(FPR) \, d(FPR)
	\end{equation}}
	\subsubsection{Label-free Metric}Moreover, we adopt the evaluation metric event structural ratio (ESR)\cite{ding2023e}. It utilizes the event contrast to evaluate the performance of denoising methods, and does not necessitate pre-labeling event sequences with valid signals or noise. ESR is composed of normalized total sum of squares (NTSS) and penalty coefficient (LN):
	\begin{equation}
		NTSS=\sum_{i=1}^{K}\frac{n_{i}(n_i-1)}{N(N-1)}
	\end{equation}
	\begin{equation}
		LN=K-\sum_{i=1}^{K}(1-\frac{M}{N})^{n_{i}}
	\end{equation}
	\begin{equation}
		ESR=\sqrt{NTSS\cdot LN}
	\end{equation}
	where $K$ is the total number of pixels in an image of warped events (IWE)\cite{Gallego2018}, $N$ is the overall number of events and $n_{i}$ represents the total events at pixel $(x_{i},y_{i})$. $M$ refers to the reference number of events used for interpolation. Detailed definitions of ESR are available in E-MLB\cite{ding2023e}.

	\begin{figure*}[htbp]
		\footnotesize
		\subfloat[1 Hz/pixel]{
			\centering 
			\begin{minipage}{0.33\linewidth} 
				\centering 
				\includegraphics[width=\linewidth]{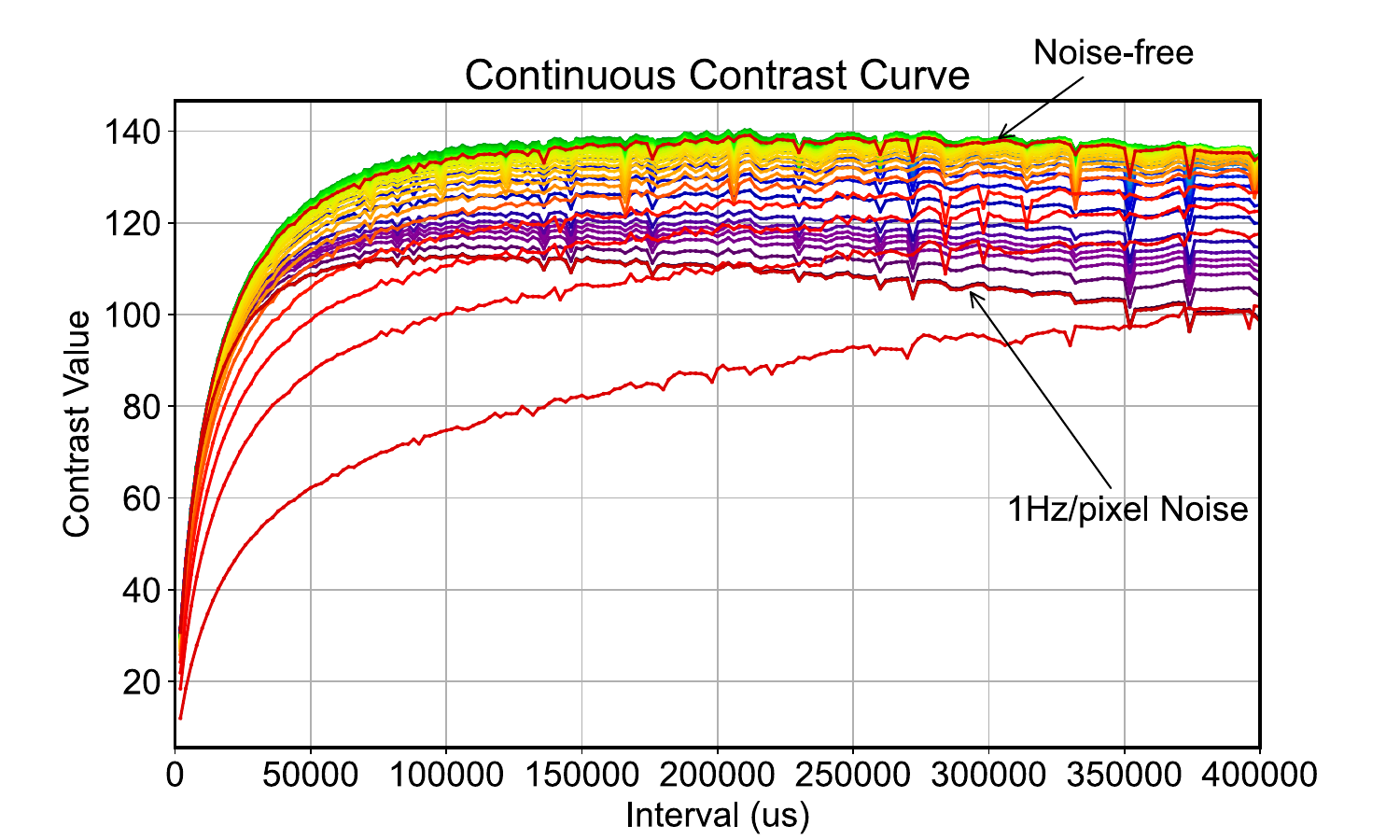}
			\end{minipage}
			\label{fig:qmlpf-1}}
		\subfloat[3 Hz/pixel]{
			\centering 
			\begin{minipage}{0.33\linewidth} 
				\centering 
				\includegraphics[width=\linewidth]{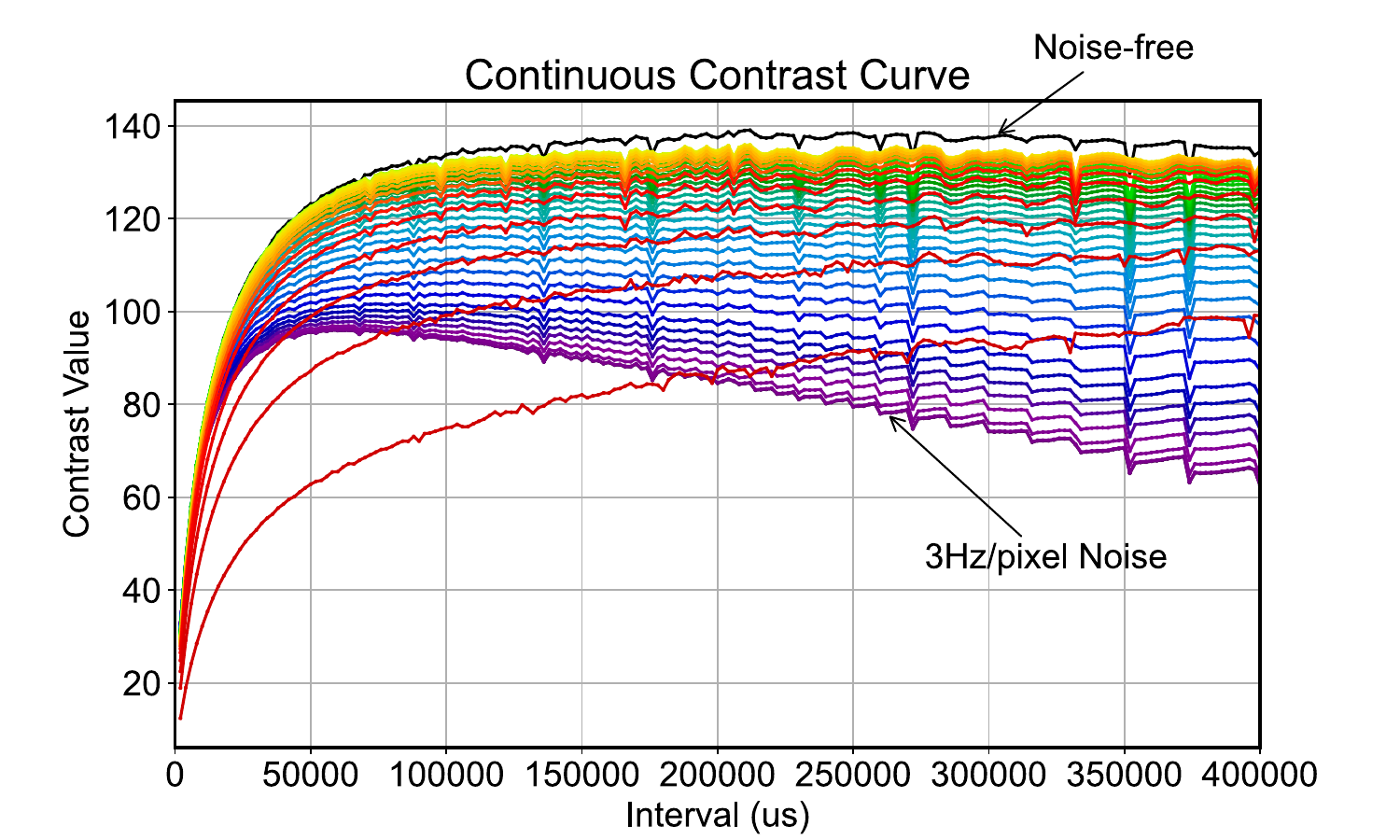}
			\end{minipage}
			\label{fig:qmlpf-2}}
		\subfloat[5 Hz/pixel]{
			\centering 
			\begin{minipage}{0.33\linewidth} 
				\centering 
				\includegraphics[width=\linewidth]{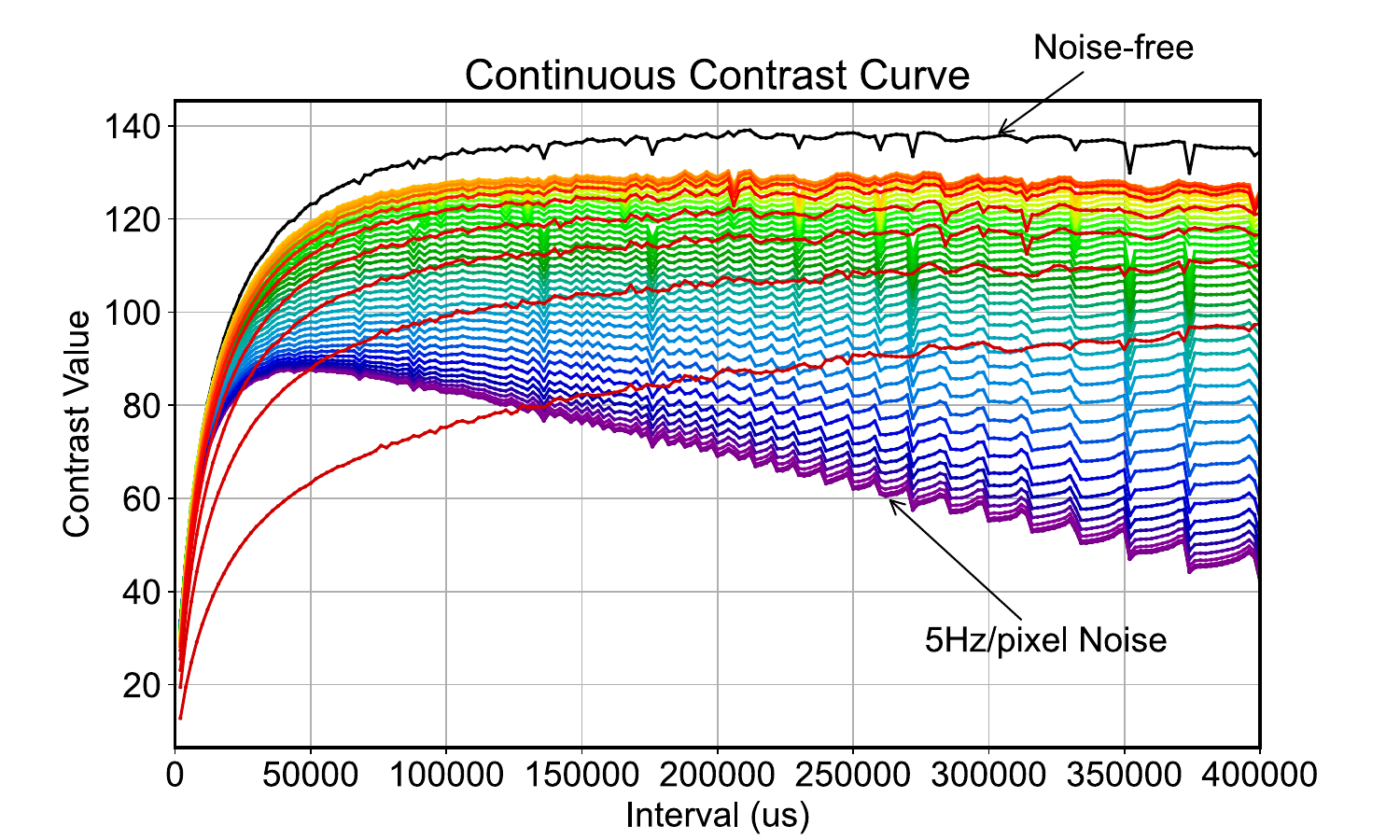}
			\end{minipage}
			\label{fig:qmlpf-3}}
		\vspace{2pt}
		\centering
		\includegraphics[width=0.8\linewidth]{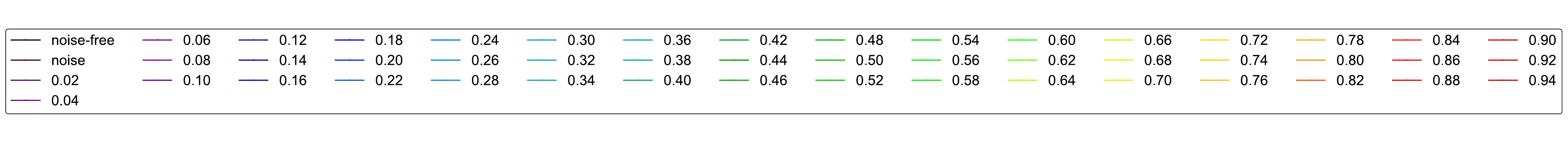}
		\caption{The continuous contrast curves of QMLPF method with varying thresholds. (a) The curves at a noise level of 1 Hz/pixel. (b) The curves at a noise level of 3 Hz/pixel. (c) The curves at a noise level of 5 Hz/pixel. In these figures, we use arrows to indicate the CCC derived directly from the unaltered event sequence and the CCC obtained from the noise-added sequence that has not been processed.}
		\label{fig:qmlpf}
	\end{figure*}
	\begin{figure*}[htbp]
		\footnotesize
		\subfloat[]{
			\centering 
			\begin{minipage}{0.33\linewidth} 
				\centering 
				\includegraphics[width=\linewidth]{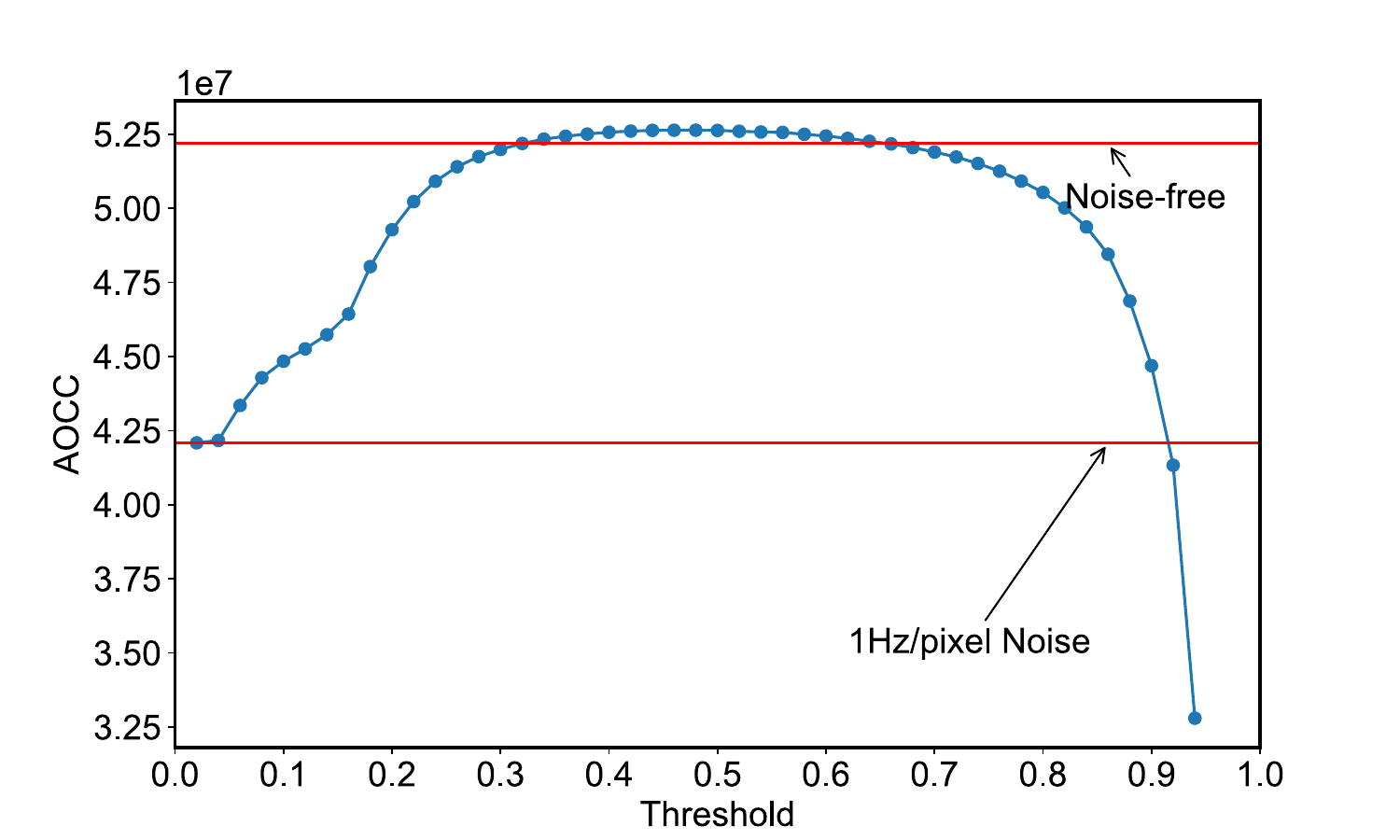}
			\end{minipage}
		}
		\subfloat[]{
			\centering 
			\begin{minipage}{0.33\linewidth} 
				\centering 
				\includegraphics[width=\linewidth]{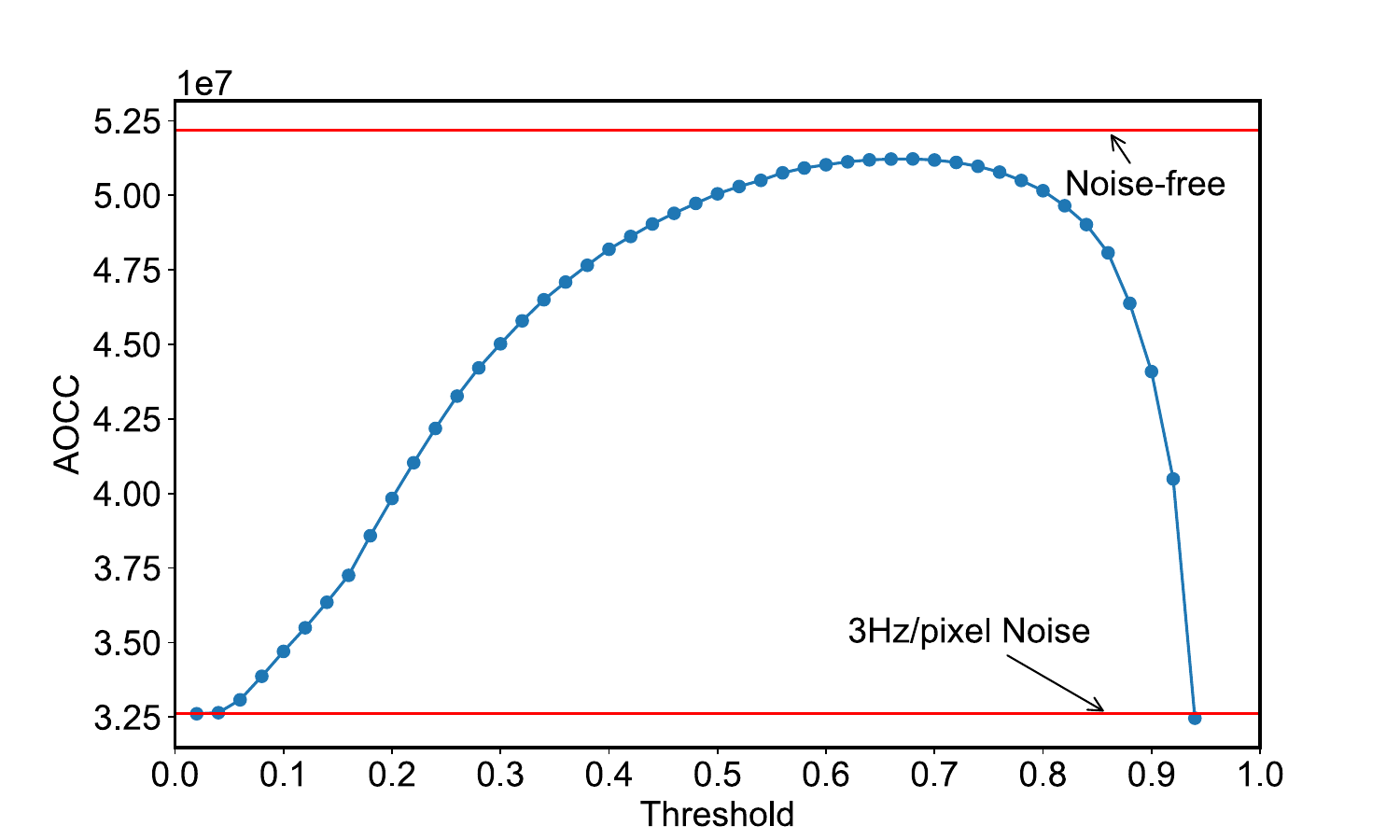}
			\end{minipage}
		}
		\subfloat[]{
			\centering 
			\begin{minipage}{0.33\linewidth} 
				\centering 
				\includegraphics[width=\linewidth]{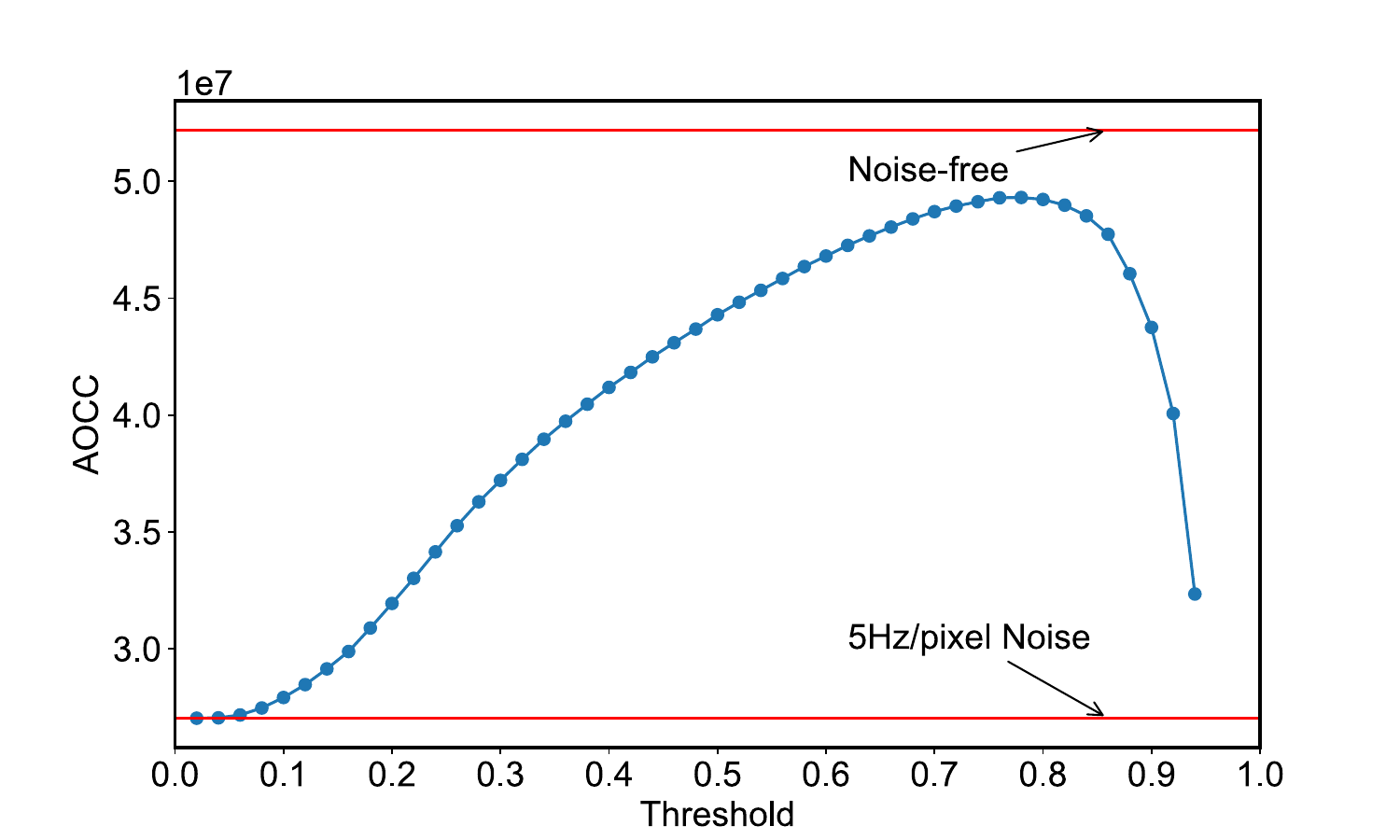}
			\end{minipage}
		}
		
		\subfloat[]{
			\centering 
			\begin{minipage}{0.33\linewidth} 
				\centering 
				\includegraphics[width=\linewidth]{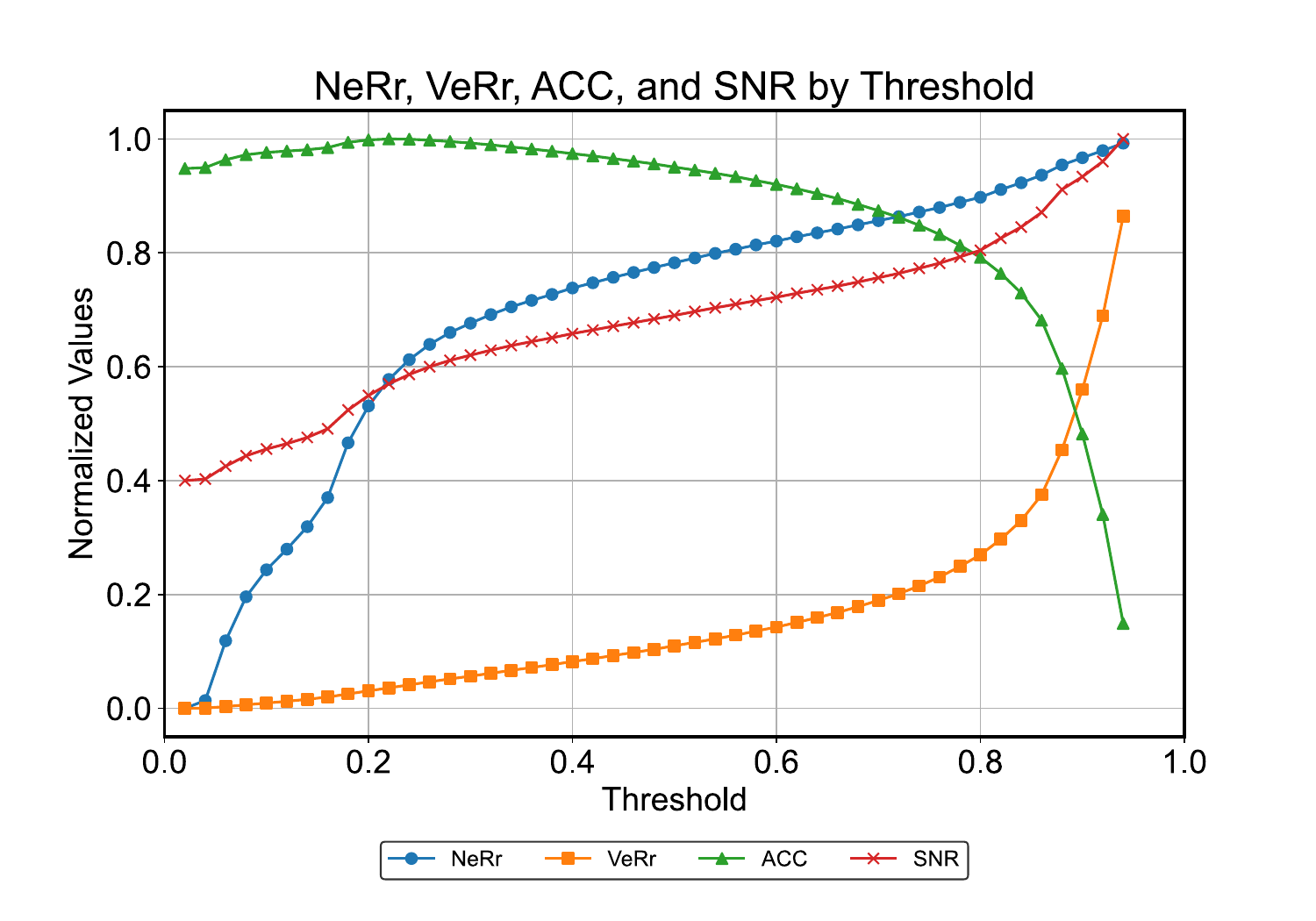}
			\end{minipage}
		}
		\subfloat[]{
			\centering 
			\begin{minipage}{0.33\linewidth} 
				\centering 
				\includegraphics[width=\linewidth]{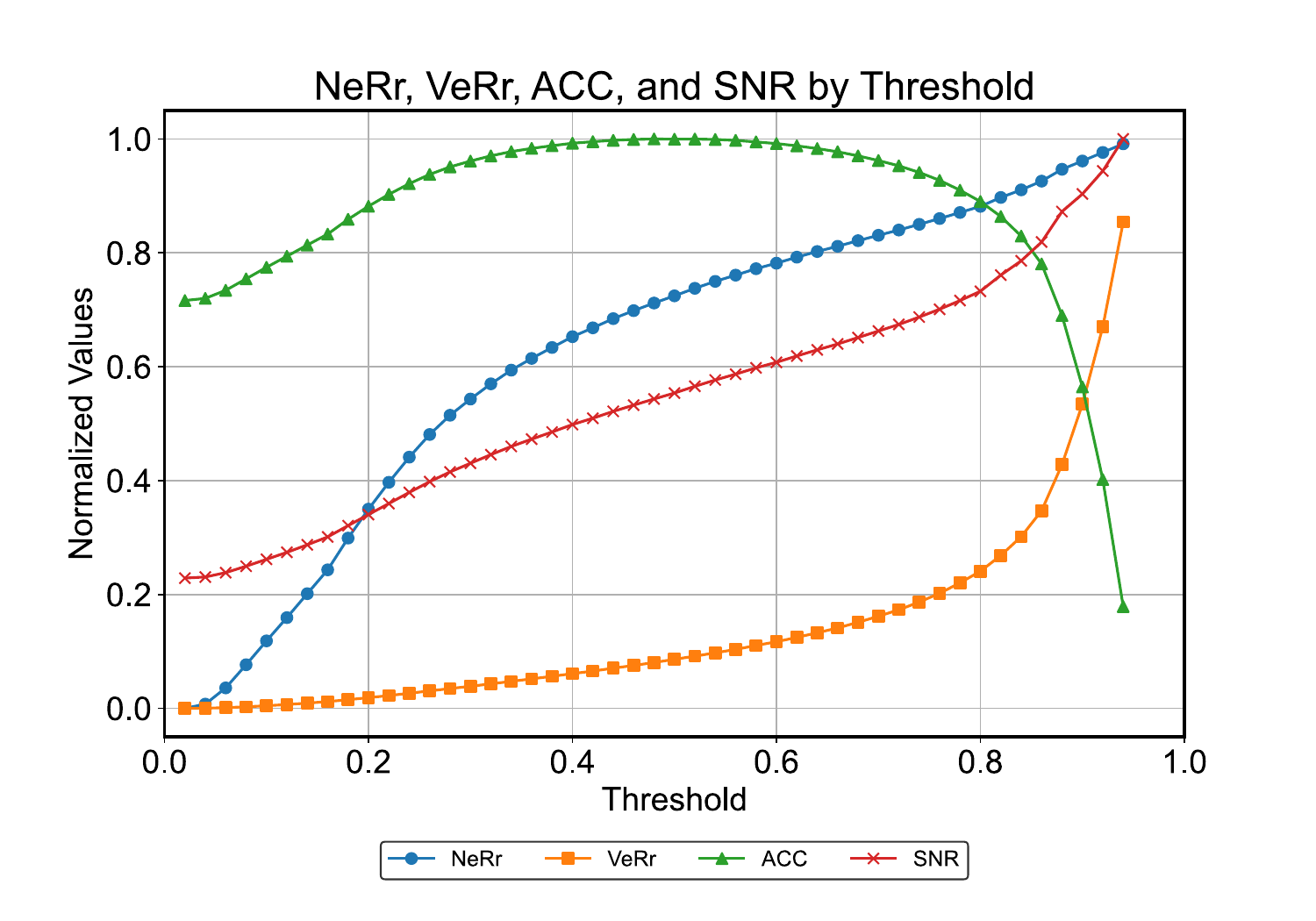}
			\end{minipage}
		}
		\subfloat[pitch]{
			\centering 
			\begin{minipage}{0.33\linewidth} 
				\centering 
				\includegraphics[width=\linewidth]{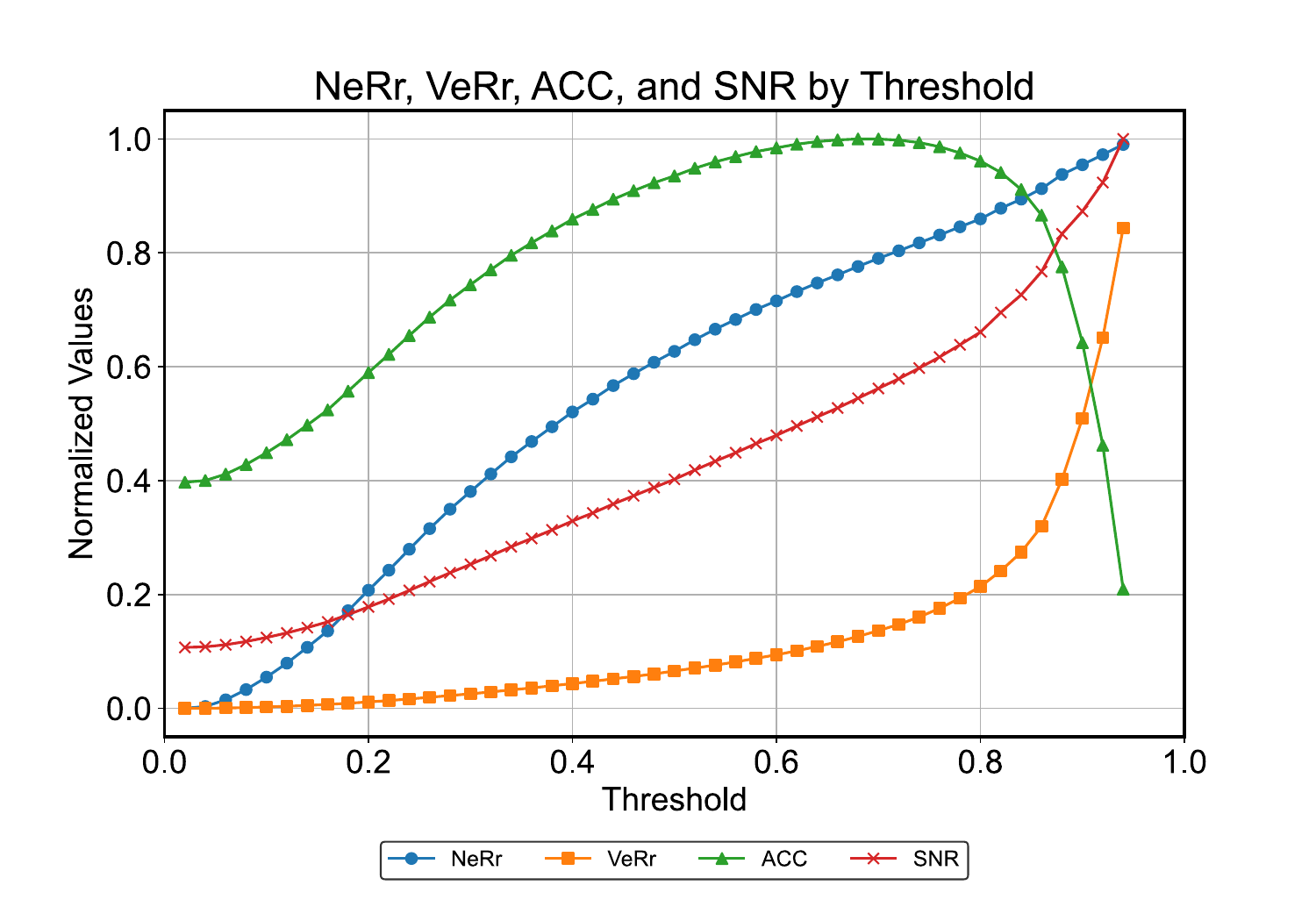}
			\end{minipage}
		}
		\caption{The AOCC, NeRr, VeRr, ACC, and SNR of QMLPF method with varying thresholds on the \textit{driving} sequence of the DND21 dataset. (a) The AOCC with a noise level of 1 Hz/pixel. (b) The AOCC with a noise level of 3 Hz/pixel. (c) The AOCC with a noise level of 5 Hz/pixel. We use red lines to indicate the AOCC values calculated for the sequence both before and after adding noise. (d) The benchmark metrics with a noise level of 1 Hz/pixel. (e) The benchmark metrics with a noise level of 3 Hz/pixel. (f) The benchmark metrics with a noise level of 5 Hz/pixel.}
		\label{fig:qmlpf_aocc}
	\end{figure*}
	\subsection{Comparison with Label-Dependent Denoising Evaluation Metrics}
	For evaluation, we utilized the QMLPF\cite{rios-navarro2023within} method. The values predicted by QMLPF method closer to 1 indicate a higher likelihood that the event is a valid signal. We then use the output from the QMLPF method under various thresholds to plot the continuous contrast curve. The threshold adjustment range for the QMLPF method spans from 0.02 to 0.98, with increments set at 0.02. Note that when the threshold exceeds 0.94, all events are removed, so results above this threshold are omitted. 
	\subsubsection{CCC of QMLPF Method}Most current denoising evaluations for event camera rely on labeled event data. We initially assess the performance of our proposed denoising metric AOCC by comparing it with existing label-dependent denoising evaluation metrics. The time interval for plotting the CCC ranges from 2 ms to 400 ms, with increments of 2 ms. When the time interval is 0, the AOCC is also equal to 0. 
	
	We plotted the continuous contrast curve for both the noise-free \textit{driving} sequence in the DND21 dataset and the noise-added \textit{driving} sequence to compare their performances. The experimental results are depicted in Fig.\ref{fig:qmlpf}

	In Fig.\ref{fig:qmlpf-1}, the QMLPF method generates different curves at various thresholds, yet they all follow the same trend. Meanwhile, it is observed that the curves for some parameters exceed those from the noise-free data. This occurs because even pure event data contains burrs. In Fig.\ref{fig:qmlpf-2} and Fig.\ref{fig:qmlpf-3}, which correspond to very high noise levels, the curves derived from both noise-free data and unprocessed noisy data envelop the data plotted under various parameters. Moreover, as the noise level increases, the curve of the denoised sequence increasingly struggles to approximate the curve of the noise-free event sequence. In other words, as the difficulty of denoising increases, the performance of denoising methods with the same parameters also decreases, which confirms the effectiveness of our proposed method.
	
	\subsubsection{AOCC vs. Label-Dependent Metrics}
	Subsequently, we present the results of AOCC, NeRr, VeRr, ACC, and SNR in Fig.\ref{fig:qmlpf_aocc}. 
	
	In the first row of Fig.\ref{fig:qmlpf_aocc}, it can be observed that the curves of AOCC are non-monotonic. However, the monotonic nature of the NeRr, VeRr, SNR curves demonstrates that these label-dependent metrics cannot be solely relied upon for evaluation, as they fail to identify an optimal solution. Moreover, ACC is a non-monotonic metric that more effectively evaluates classification outcomes. However, the accuracy of classification and the quality of the denoising effect are not synonymous. For instance, in the 1 Hz/pixel sequence, both low NeRr and VeRr also attained high ACC values. 
	
\begin{figure}[htbp]
		\centering
		\includegraphics[width=0.7\linewidth]{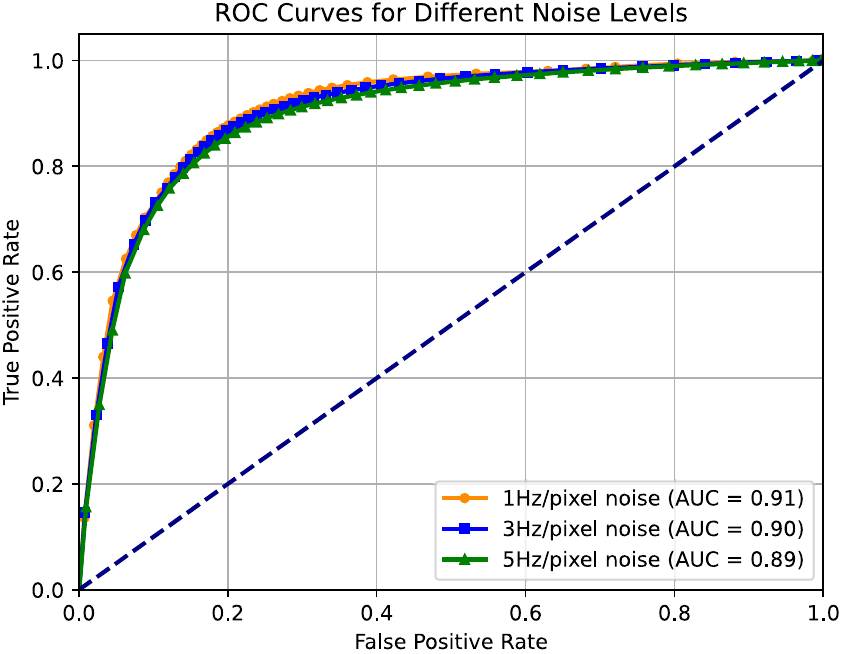}
		\caption{ROC and AUC of QMLPF method under different noise levels.}
		\label{fig:roc}
	\end{figure}

	Next, we demonstrate the ROC curve of the QMLPF method under different noise levels and calculate the AUC, as shown in Fig.\ref{fig:roc}. ROC, which is plotted using the TPR and FPR of the denoising method under different parameters, provides a more objective evaluation. However, it still relies on labeled data and is not convenient for selecting the optimal parameters.
	\begin{figure*}[htbp]
		\centering
		\subfloat[1 Hz/pixel]{
			\begin{minipage}{0.33\linewidth}
				\centering
				\includegraphics[width=\linewidth]{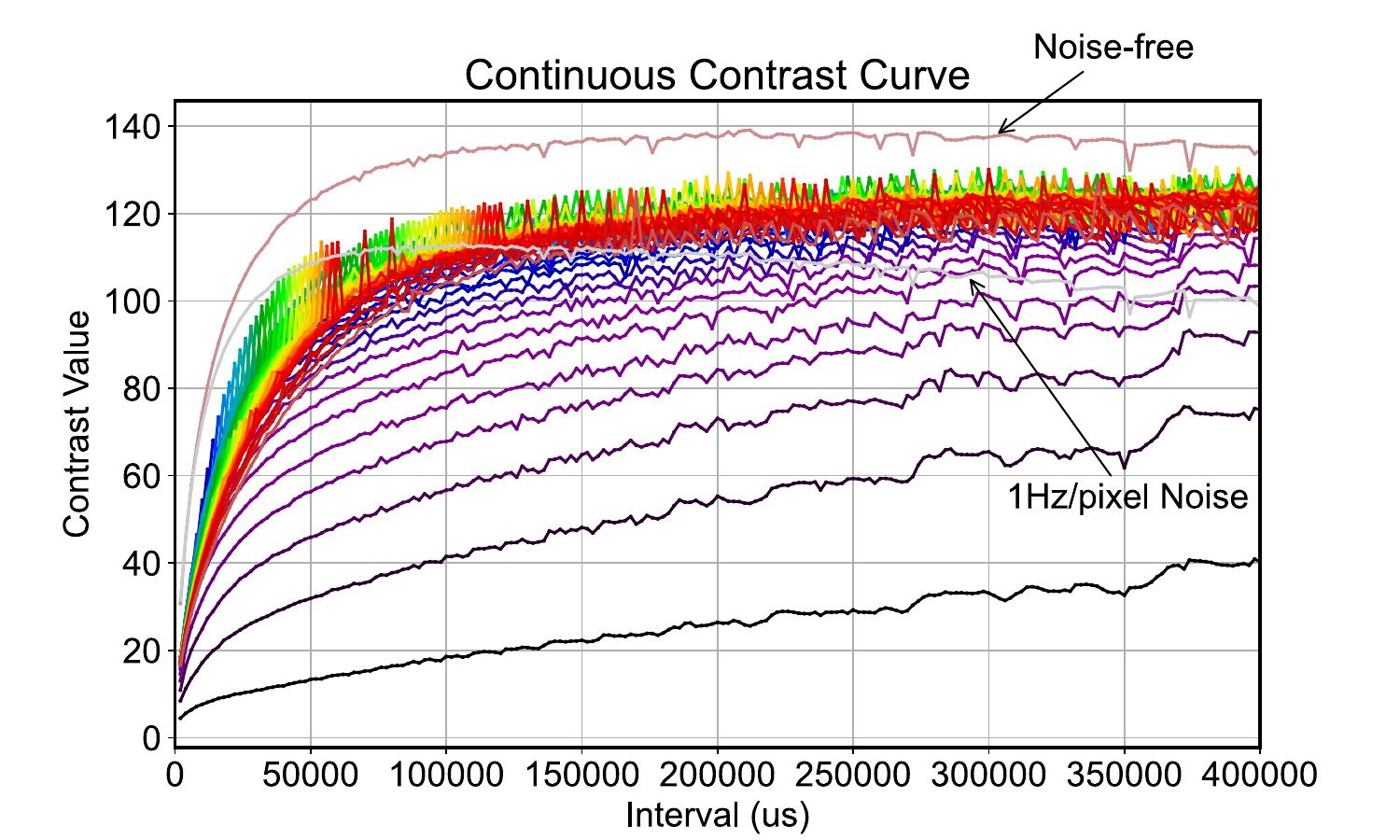}
		\end{minipage}}
		\subfloat[3 Hz/pixel]{
			\begin{minipage}{0.33\linewidth}
				\centering
				\includegraphics[width=\linewidth]{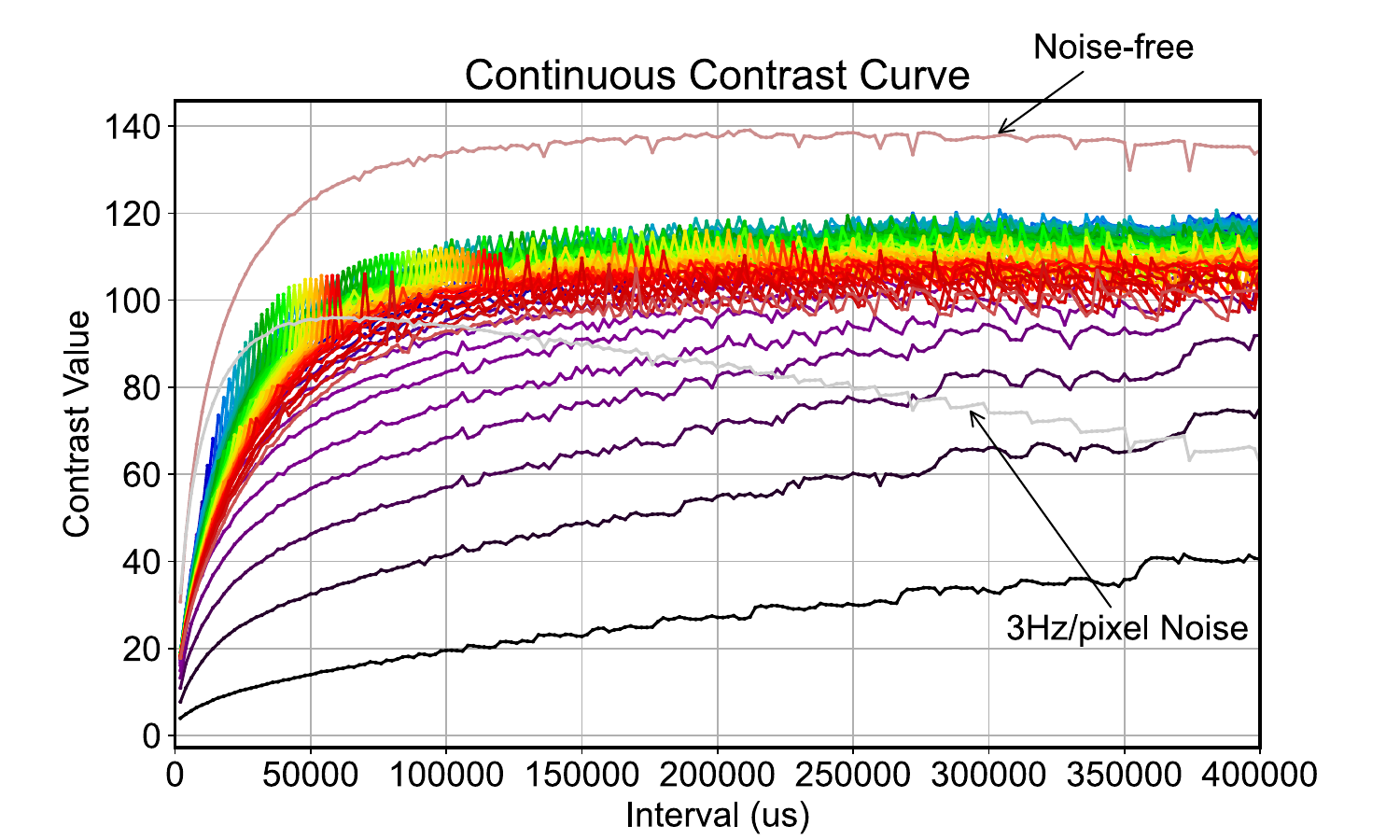}
		\end{minipage}}
		\subfloat[5 Hz/pixel]{
			\begin{minipage}{0.33\linewidth}
				\centering
				\includegraphics[width=\linewidth]{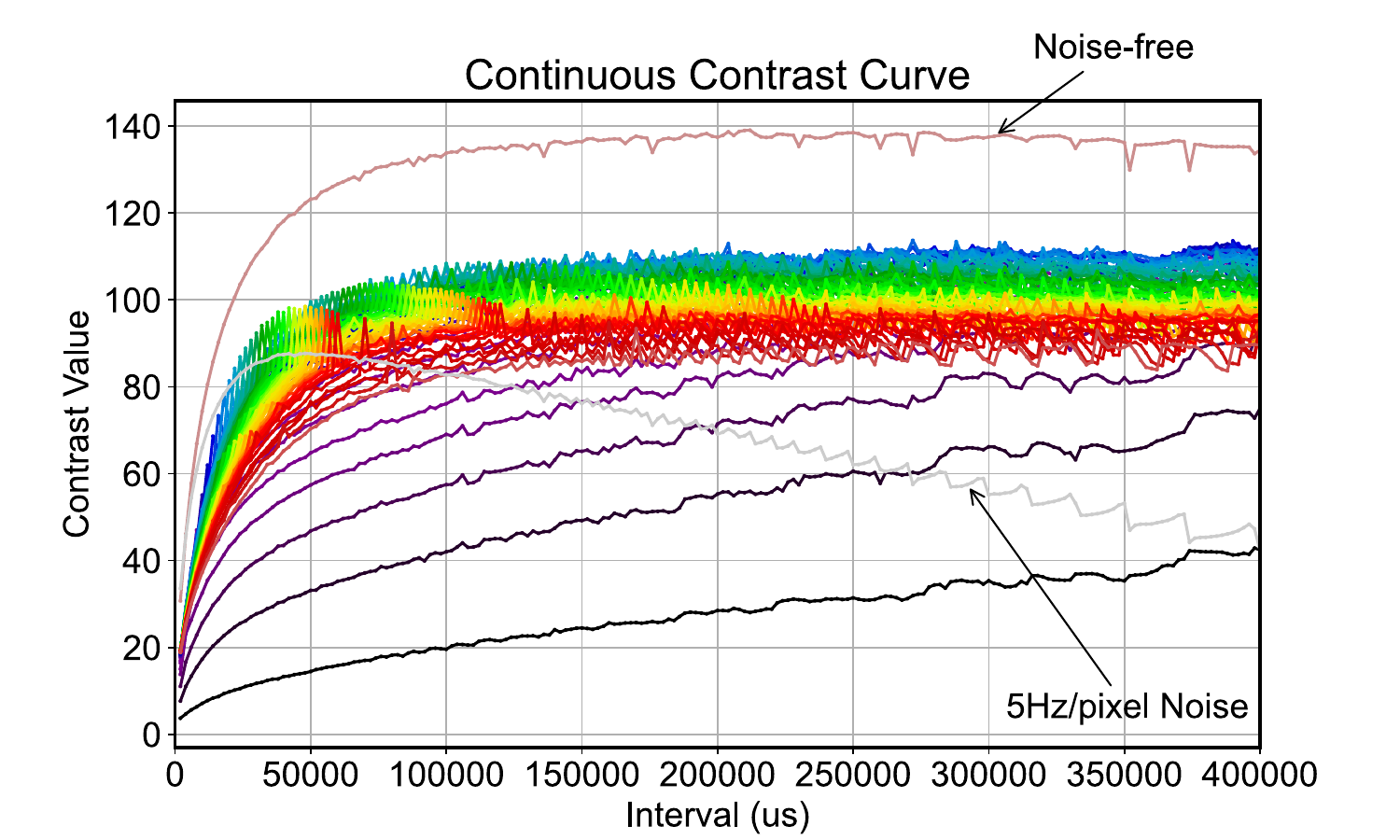}
		\end{minipage}}
		\vspace{2pt}
		\centering
		\includegraphics[width=0.8\linewidth]{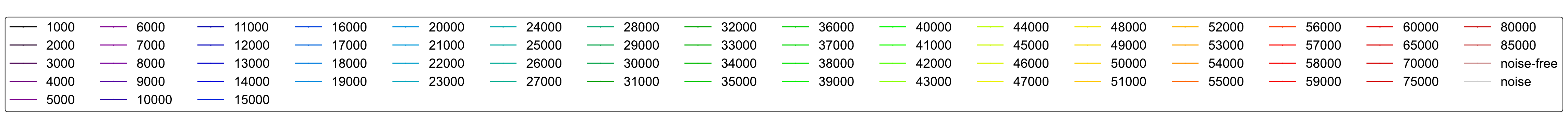}
		
		\subfloat[]{
			\begin{minipage}{0.33\linewidth}
				\centering
				\includegraphics[width=\linewidth]{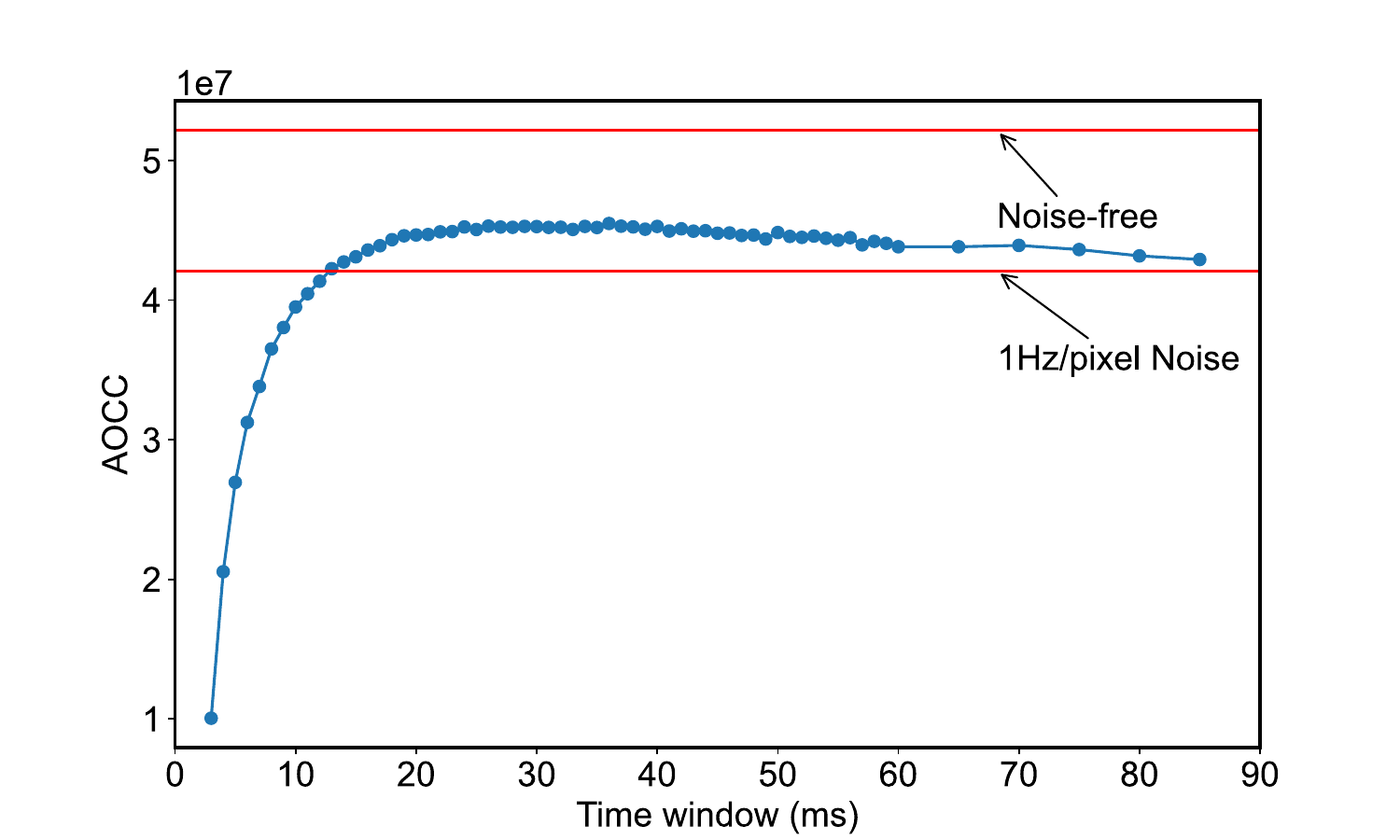}
		\end{minipage}}
		\subfloat[]{
			\begin{minipage}{0.33\linewidth}
				\centering
				\includegraphics[width=\linewidth]{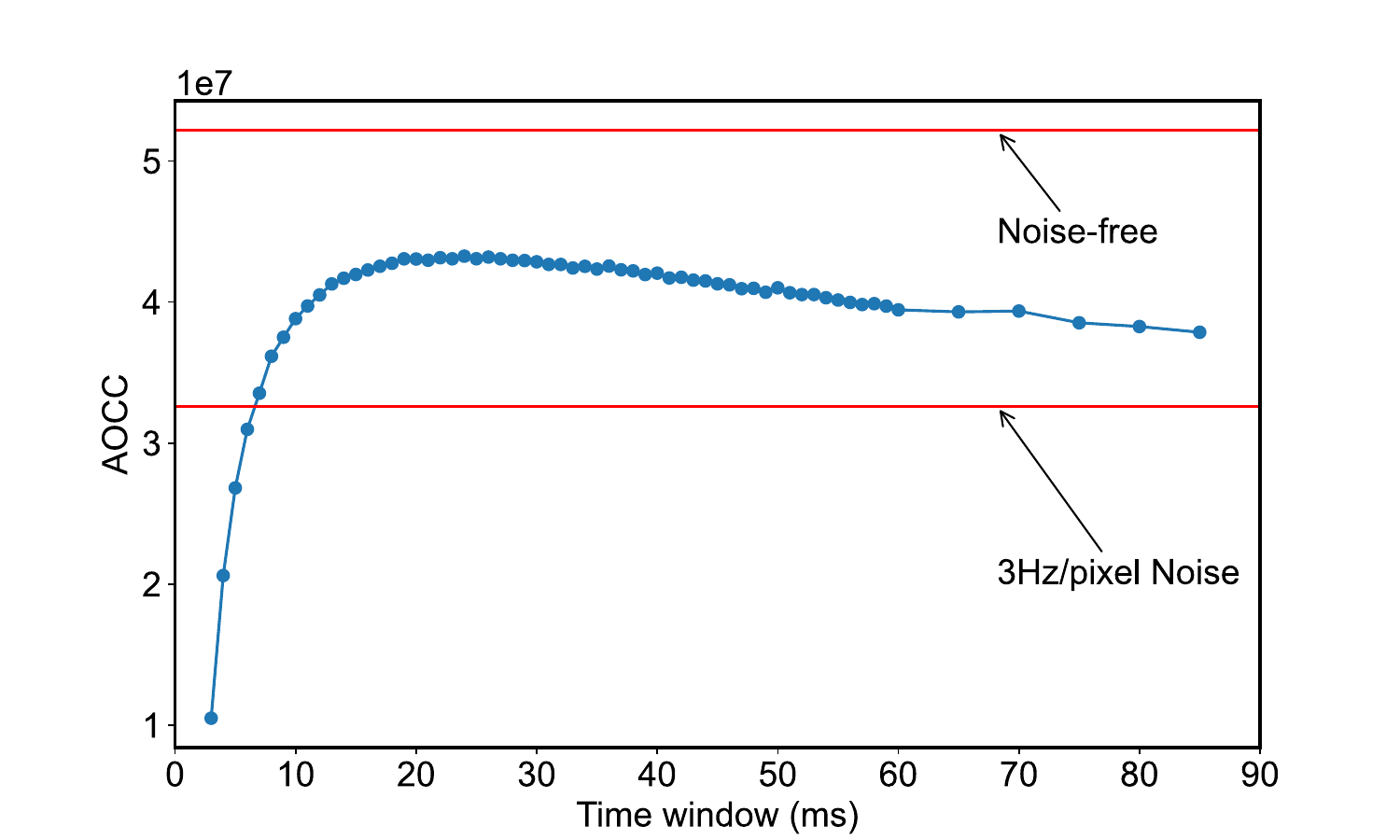}
		\end{minipage}}
		\subfloat[]{
			\begin{minipage}{0.33\linewidth}
				\centering
				\includegraphics[width=\linewidth]{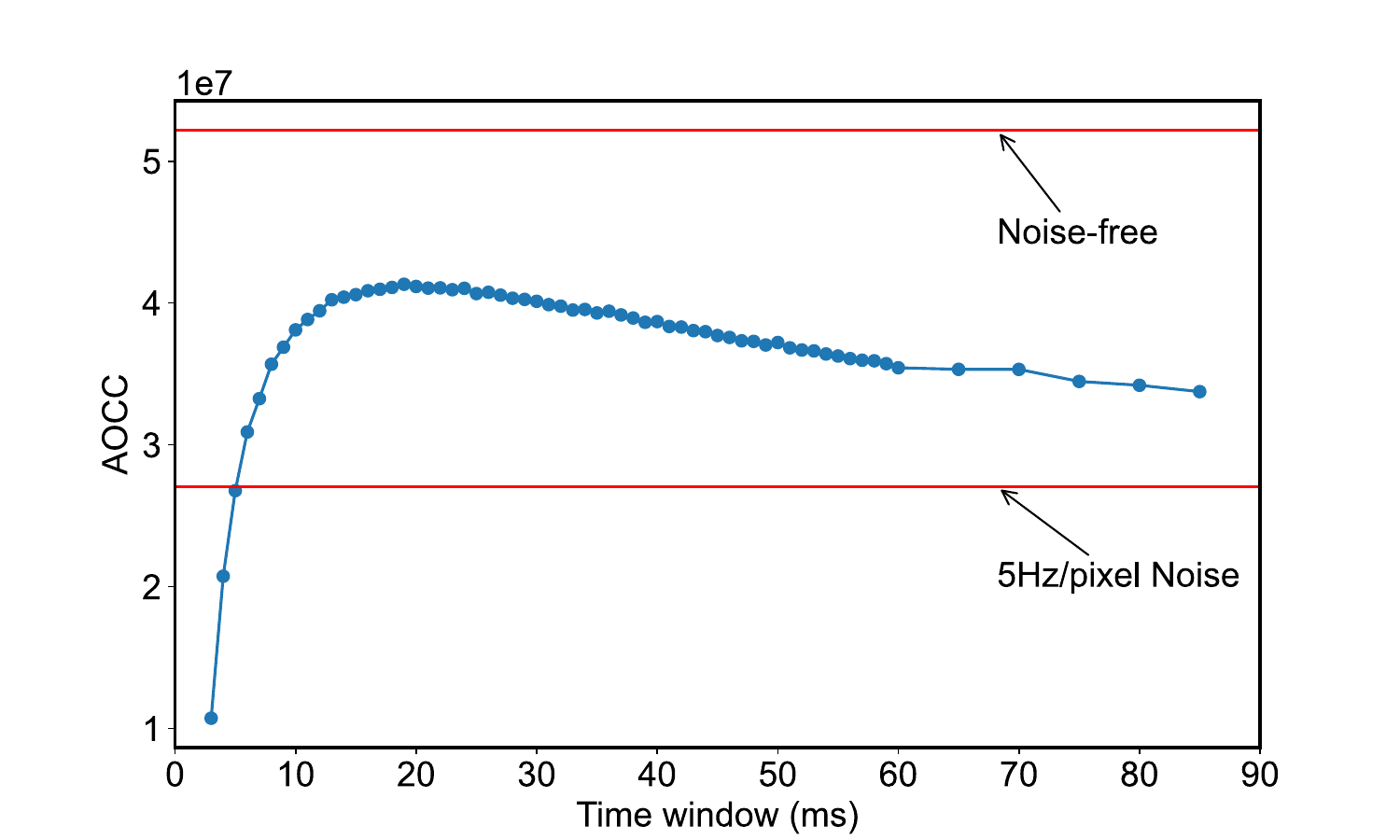}
		\end{minipage}}
		
		\caption{CCC and AOCC of EventZoom method on \textit{driving} sequence of the DND21 dataset with different time windows. (a) The CCC with a noise level of 1 Hz/pixel. (b) The CCC with a noise level of 3 Hz/pixel. (c) The CCC with a noise level of 5 Hz/pixel. (d) The AOCC with a noise level of 1 Hz/pixel. (e) The AOCC with a noise level of 3 Hz/pixel. (f) The AOCC with a noise level of 5 Hz/pixel.}
		\label{fig:eventzoom}
	\end{figure*}
	\subsubsection{AOCC for Optimal Parameter Selection}
	Additionally, the AOCC we proposed can easily identify the optimal parameters for the denoising method, or the interval where these optimal parameters are located, which is highly beneficial for fair comparisons.
	
	We further outline the NeRr, VeRr, SNR, and ACC values computed at the threshold parameters that yield the optimal AOCC in Table.\ref{tab:1}.
	
	\begin{table}[htbp]
		\centering
		\begin{threeparttable}
			\caption{The best AOCC metric with other label-based metric.}
			\label{tab:1}
			\renewcommand{\arraystretch}{1.3} 
			\setlength{\tabcolsep}{2pt}
			\begin{tabular}{lccccccc}
				\hline
				\toprule
				Method& Noise level&AOCC(1e7)&NeRr&VeRr&ACC&SNR&Threshold\\
				\midrule
				\multirow{3}{*}{QMLPF}&1 Hz/pixel	&5.26	&0.77	&0.10	&0.88	&14.42	&0.48\\
				&3 Hz/pixel	&5.12	&0.82	&0.15	&0.84	&10.25	&0.68\\
				&5 Hz/pixel	&4.93	&0.85	&0.19	&0.82	&8.93	&0.78\\
				\bottomrule
				\hline
			\end{tabular}
		\end{threeparttable}
		
	\end{table}
	
	In Table.\ref{tab:1}, attaining the optimal AOCC correlates with a balance of NeRr, VeRr, ACC, and SNR. When the threshold is maximized, the AOCC achieved is substantially lower, characterized by a markedly high VeRr and NeRr, a significantly reduced ACC, and the maximum SNR. This scenario indicates that while a considerable amount of noise is eliminated, the residual effective signal is minimal, suggesting an inadequate denoising outcome despite the elevated SNR. Conversely, lowering the threshold results in a decreased AOCC, reductions in both VeRr and NeRr, a lower SNR, and a significantly high ACC. Although this configuration retains a substantial amount of effective signals, it removes minimal noise, indicating a suboptimal denoising effect. Therefore, AOCC serves as a non-monotonic index that provides an objective measure of denoising efficacy.

	\subsection{Comparison with Label-Free Denoising Evaluation Metric}
	\subsubsection{Evaluation on Label-Free Denoising Method}
	
	Some denoising methods project events, altering their coordinates and quantities and rendering pre-marked labels ineffective, typically evaluated indirectly through downstream tasks, whereas our proposed method allows for direct evaluation.
	
	We choose EventZoom \cite{duan2021eventzoom} method for our evaluation. EventZoom is a composite, learning-based method that uses event sequences to generate super-resolution images, with event denoising serving as a secondary benefit of the super-resolution task. We utilize the aforementioned consistent \textit{driving} sequence to evaluate EventZoom, which adjusts time windows from 1 ms to 60 ms with a step of 1 ms for accumulating event frames to achieve diverse denoising effects. For intervals exceeding 60 ms, we use increments of 5 ms up to 85 ms to simplify the calculations. The experimental results are shown in Fig.\ref{fig:eventzoom}.

	In Fig.\ref{fig:eventzoom}, the AOCC curve is also non-monotonic. Using the AOCC curve, the optimal parameters for the EventZoom method can be easily determined, facilitating comparison with other denoising methods. Moreover, it is observed that EventZoom is relatively insensitive to the selection of time interval parameters, performing consistently when parameters are chosen between 15 ms and 30 ms. However, EventZoom exhibits the problem of over-denoising, as the peak value of the AOCC curve is significantly lower than that of the sequence without noise.
	
	\subsubsection{AOCC vs. Label-Free Metric}
	We selected the recently published label-free event-based denoising evaluation metric ESR\cite{ding2023e} for comparison. We adopt E-MLB\cite{ding2023e} dataset, which corresponds to the ESR, as benchmark dataset. The E-MLB dataset, recorded using a DAVIS346 event camera, inherently contains noise. We selected the \textit{Nighttime-Lounge-ND00-1} sequence for our analysis. This sequence contains a significant amount of noise.
	
\begin{figure*}[htbp]
		\footnotesize
		\centering 
		\subfloat[]{
			\centering 
			\begin{minipage}{0.19\linewidth} 
				\centering 
				\includegraphics[width=\linewidth]{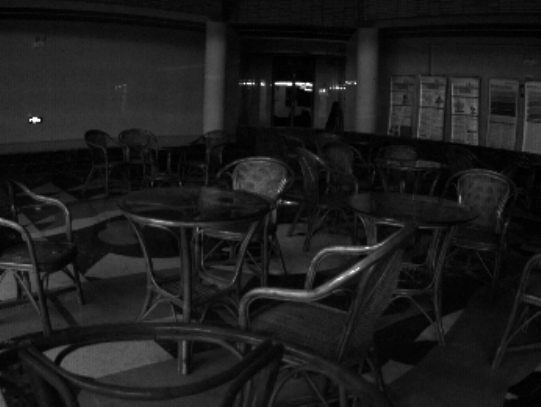}
			\end{minipage}
			\label{fig:dwf1}
		}
		\subfloat[]{
			\centering 
			\begin{minipage}{0.19\linewidth} 
				\centering 
				\includegraphics[width=\linewidth]{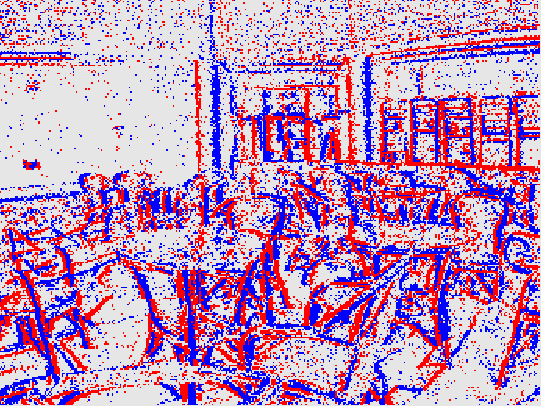}
			\end{minipage}
			\label{fig:dwf2}
		}
		\subfloat[]{
			\centering 
			\begin{minipage}{0.19\linewidth} 
				\centering 
				\includegraphics[width=\linewidth]{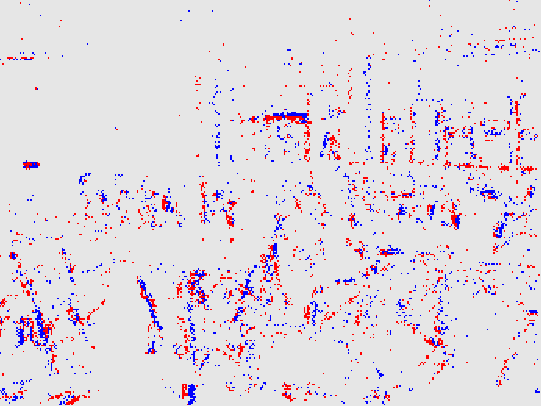}
			\end{minipage}
			\label{fig:dwf3}
		}
		\subfloat[]{
			\centering 
			\begin{minipage}{0.19\linewidth} 
				\centering 
				\includegraphics[width=\linewidth]{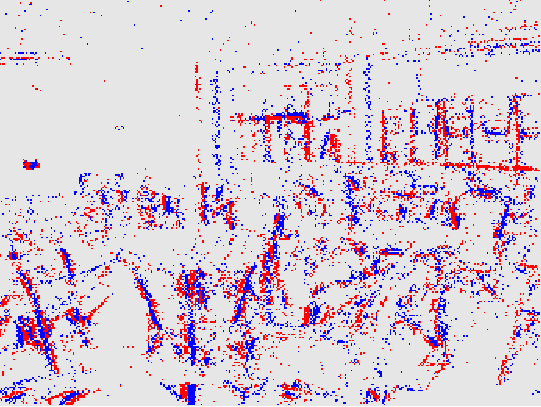}
			\end{minipage}
			\label{fig:dwf4}
		}
		\subfloat[]{
			\centering 
			\begin{minipage}{0.19\linewidth} 
				\centering 
				\includegraphics[width=\linewidth]{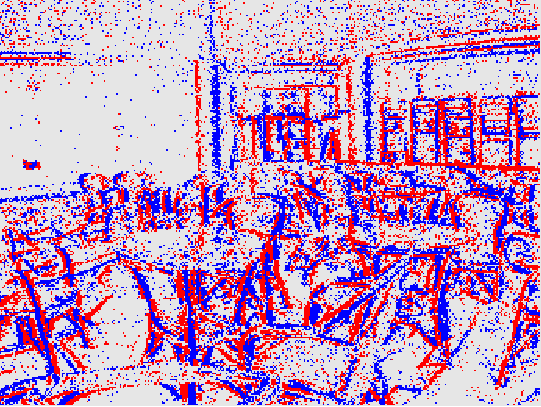}
			\end{minipage}
			\label{fig:dwf5}
		}
		\caption{(a) The RGB frame captured by the DAVIS346. (b) The event frame in \textit{lounge} sequence corresponding to (a) that has not been denoised. (c) The event frame generated by DWF with a buffer size of 200 and search radius of 2. (d) The event frame generated by DWF with a buffer size of 200 and search radius of 4. (e) The event frame generated by DWF with a buffer size of 200 and search radius of 14.}
		\label{fig:dwf}
	\end{figure*}

\begin{figure}[htbp]
		\footnotesize
		\centering 
		\includegraphics[width=0.7\linewidth]{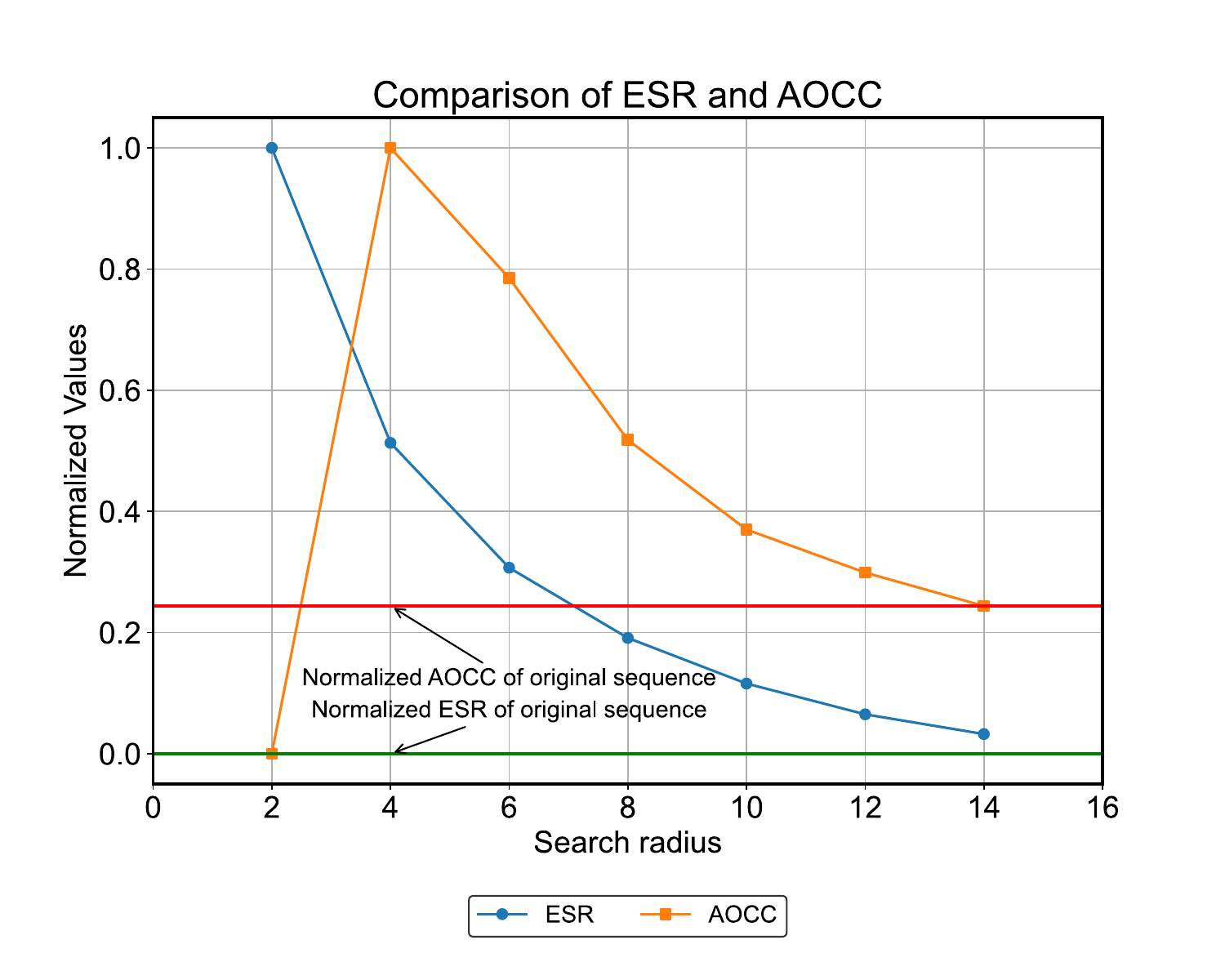}
		\caption{The AOCC and ESR values adjusted by DWF method parameters. The results of ESR are monotonic, the more valid signals and noise are removed simultaneously, the higher the score. This scoring mechanism contradicts the goal of the denoising task, which is to retain more effective signals while removing more noise. In contrast, our AOCC metric objectively evaluates the denoising performance of DWF under different parameters.}
		\label{fig:dwf-a}
	\end{figure}

	The DWF \cite{guo2022low} method is used for evaluation which includes two main adjustable parameters: search radius and buffer size. Decreasing the search radius and shortening the buffer size significantly increases the removal of both noise and valid signals. We vary the search radius from 2 to 14 in steps of 2, keeping the buffer size constant at 200, to explore different denoising effects. The visual demonstrations are shown in Fig.\ref{fig:dwf}.  The AOCC and ESR values for sequences processed by the DWF method using various parameters are shown in Fig.\ref{fig:dwf-a}. 

	In Fig.\ref{fig:dwf-a}, the ESR scores indicate that the highest scores are achieved when the the search radius is the smallest. However, similar to SNR, more aggressive denoising tends to yield higher scores. Fig.\ref{fig:dwf2} demonstrates that the events are primarily triggered by the lounge scene, with events largely consisting of noise. Fig.\ref{fig:dwf3} displays the event frame produced by the DWF method, which achieved the highest ESR score. However, the AOCC yielded the lowest score. Although it effectively removes most of the noise, it also erroneously eliminates a significant number of valid events. Fig.\ref{fig:dwf4}, achieving a lower ESR score, while retains more valid events and still effectively reduces noise. Conversely, the AOCC awarded higher scores. In Fig.\ref{fig:dwf5}, due to the extremely lenient criteria for denoising, almost no noise events are removed. Therefore, the AOCC and ESR scores of it are low. 
	
	In summary, these experimental results suggest that the ESR curve is monotonic, while the AOCC curve is non-monotonic. This underscores the effectiveness and objectivity of our proposed AOCC.
	
	\section{Discussions}
	\label{sec7}
	
	Denoising in event cameras is a fundamental task that is essential for preserving their high efficiency and low power consumption. Current metrics that depend on labeled data face significant challenges in practical scenarios, primarily because the vast majority of event data from event camera is captured directly without labels. While using downstream tasks for evaluation is a feasible approach, the diversity of such tasks means that this indirect method may not comprehensively assess denoising performance. Moreover, existing label-free evaluation metrics are often monotonic, leading to erroneous assessments. They tend to reward the removal of both noise and effective signals, which contradicts the primary objective of denoising: to retain as much effective signal as possible while eliminating noise.
	
	Although our proposed metric AOCC is highly effective in evaluating the performance of denoising methods, it does have some limitations, primarily because it requires traversing multiple time intervals, leading to more computational cost. However, as an evaluation metric for denoising, computing speed is not its primary objective. Despite the slower calculations, AOCC offers distinct advantages as a label-free and non-monotonic metric for evaluating denoising of event camera.
	
	\section{Conclusions}
	\label{sec8}
	In this paper, we introduce AOCC, a non-monotonic, label-free evaluation metric designed specifically for denoising in event cameras. This metric effectively discriminates between denoising methods, awarding higher scores to those that achieve a high rate of noise removal while minimally impacting effective signals. Conversely, methods that perform poorly in denoising receive punitive scores. Unlike label-dependent metrics, which lack a unified standard and often necessitate a multifaceted evaluation approach, AOCC provides a singular, standardized measure of performance. Additionally, its independence from labeled data enhances its applicability across various event camera deployments, making it a robust tool for evaluating denoising efficacy in real-world scenarios.

	\bibliographystyle{IEEEtran}
	\bibliography{IEEEabrv, rotation}
	\vfill
	
\end{document}